\documentclass[lettersize,journal]{IEEEtran}
\usepackage{amsmath,amsfonts}
\usepackage{algorithmic}
\usepackage{array}
\usepackage[caption=false,font=normalsize,labelfont=sf,textfont=sf]{subfig}
\usepackage{textcomp}
\usepackage{stfloats}
\usepackage{url}
\usepackage{verbatim}
\usepackage{graphicx}
\newtheorem{theorem}{Theorem}
\usepackage{algorithm}
\usepackage{booktabs}
\usepackage{multirow}
\usepackage{xcolor}
\usepackage{color}
\usepackage{diagbox}
\definecolor{gre}{RGB}{0, 205, 0}
\def\BibTeX{{\rm B\kern-.05em{\sc i\kern-.025em b}\kern-.08em
    T\kern-.1667em\lower.7ex\hbox{E}\kern-.125emX}}
\usepackage{balance}
\begin{document}
\title{Modality Equilibrium Matters: Minor-Modality-Aware Adaptive Alternating for
Cross-Modal Memory Enhancement}

\author{Xiang~Shi$^\dagger$, Rui~Zhang$^\dagger$,~\IEEEmembership{Member,~IEEE,}
    Jiawei~Liu, Yinpeng~Liu, Zhu Liang, Qikai~Cheng$^*$,
    and~Wei~Lu$^*$ 
   \thanks{$\dagger$ denotes the equal contribution.}
   \thanks{$*$ denotes the corresponding author.}
    \thanks{All authors are with the School of Information Management, Wuhan University, Wuhan 430072, Hubei, P. R. China.}}

\maketitle

\begin{abstract}
Multimodal fusion is susceptible to modality imbalance, where dominant modalities overshadow weak ones, easily leading to biased learning and suboptimal fusion, especially for incomplete modality conditions. To address this problem, we introduce an Equilibrium Deviation Metric (EDM) to quantify this imbalance and verify, in both theoretical and empirical terms, that the optimization order of modalities plays a critical role in approaching equilibrium. In particular, we demonstrate that an EDM-ranked weak-to-strong schedule achieves the tightest convergence bound among all possible ordering strategies. Leveraging these insights, we design an alternating strategy that dynamically prioritises under-optimised modalities, plus a modality-mapping layer for feature alignment and a memory module for information filtering and inheritance. Our framework is compatible with both conventional and MLLM-based backbones. It achieves new state-of-the-art (SOTA) on four benchmarks (e.g., +3.36\% on CREMA-D, +3.51\% on Kinetics-400), and remains robust under missing-modality conditions. These findings highlight the value of modality scheduling, offering a principled alternative to conventional joint training.
\end{abstract}


\section{Introduction}
\IEEEPARstart{M}{ultimodal} learning has witnessed remarkable progress with the advent of multimodal large language models (MLLMs) \cite{Yin_2024}, which enable unified understanding and reasoning across diverse modalities such as vision, language, and audio. By leveraging cross-modal interactions, these models boost performance across a wide range of tasks. However, a fundamental challenge in multimodal learning, i.e., \textbf{modality laziness} or \textbf{modality bias} is still ubiquitous, where dominant modalities tend to overshadow weaker ones, leading to insufficient optimization. This phenomenon is analogous to the “Buckets effect”, where the weakest modality limits the overall learning capacity, imposing a lower bound on joint optimization when modality capabilities are imbalanced.

To address modality laziness, various training paradigms have been explored to balance unimodal and multimodal learning objectives. Broadly speaking, these can be categorized into \textbf{joint training} and \textbf{alternating training}.

\textbf{Joint training} optimizes all modalities simultaneously by enforcing cross-modal interactions throughout learning \cite{peng2022balanced,liu2023visual}. While this approach is simple and efficient, it often overlooks the development of strong unimodal representations. As a result, models trained jointly may perform poorly when confronted with missing or corrupted modalities, revealing their limited robustness in real-world multimodal scenarios.

\textbf{Alternating training}, in contrast, optimizes each modality independently in a stepwise manner \cite{gunes2005affect, zhang2024multimodal}, encouraging more robust unimodal capabilities. However, as pointed by \cite{hua2024reconboost}, unimodal competence alone is insufficient. Without effective alignment and fusion mechanisms, alternating strategies may fail to fully exploit cross-modal complementarity. Therefore, a robust multimodal training paradigm must not only preserve unimodal strengths but also promote balanced cross-modal integration, ensuring both adaptability and generalization across diverse multimodal settings.

To address the challenges of modality imbalance and under-optimization, our study centers on alternating learning, with particular attention to how the update order of modalities affects optimization. We propose an equilibrium evaluation metric, \textbf{Equilibrium Deviation Metric (EDM)}, to quantify imbalance during training. Guided by EDM, we show that the update sequence is critical for convergence. In particular, a weak-to-strong order achieves the tightest convergence bound and consistently outperforms strong-to-weak or random schedules. Based on these insights, we develop an alternating training framework that dynamically adjusts the training sequence according to EDM scores. To further promote integration, we introduce a cross-modal alignment module that includes a modality mapping mechanism for aligning feature distributions and a modality-aware memory module for retaining critical information across training stages. Our framework is compatible with various encoder backbones and achieves superior results even when applied to MLLMs.

Our main contributions are summarized as follows:
\begin{itemize}
    \item We introduce an \textbf{Equilibrium Deviation Metric} to quantify modality imbalance during training and guide modality scheduling. It supports our key theoretical insight that optimizing the update order in alternating learning is essential for driving the model toward modality equilibrium.
    \item We propose an \textbf{EDM-guided alternating training framework} that dynamically prioritizes under-optimized modalities, enabling balanced updates and alleviating modality laziness throughout the training process.
    \item We design a \textbf{modality-aware cross-modal memory module} that promotes structural alignment across modalities while filtering and inheriting key information. Extensive experiments across four benchmarks demonstrate our method’s superior performance and robustness under missing-modality conditions.
     
\end{itemize}   

\section{Related Work}

\label{sec:related}

\subsection{Imbalanced Problem in Multimodal Learning}

\noindent Multimodal learning involves integrating information from different modalities, such as text, images, and audio, to enhance model performance in diverse real-world applications. However, not all modalities contribute equally in every scenario. For example, in visual question answering (VQA), both vision and language are crucial, whereas in multimodal sentiment analysis, facial expressions often outweigh speech in determining sentiment. This imbalance extends to multimodal deep learning, where models often develop \textbf{ modality laziness} or \textbf{modality bias}, over-relying on dominant modalities while under-utilizing weaker ones.

Such bias leads to poor generalization in missing-modality scenarios and increased vulnerability to adversarial attacks \cite{vishwamitra2021understanding, yang2024quantifying}. A well-trained model may perform robustly when all modalities are present but degrade significantly when one is absent. This phenomenon is common in naive joint training and fine-tuning of MLLMs. Studies suggest two key reasons for this issue: (1) modalities overfit and generalize at different rates \cite{wang2020makes,sun2021learning}, with dominant modalities suppressing weaker ones; (2) fusion architectures reinforce unimodal dominance, as late and intermediate fusion introduce prolonged unimodal learning phases, leading to permanent bias \cite{zhangunderstanding}.

\subsection{Multimodal Learning Strategies}

\noindent Existing multimodal learning strategies primarily focus on \textbf{joint training}, where all modality-specific encoders are optimized simultaneously. To enhance this approach, one line of research focuses on adaptive optimization, dynamically adjusting modality-specific gradients to balance learning contributions \cite{peng2022balanced}. Another direction investigates fusion strategies, where uncertainty estimation techniques are employed to improve multimodal robustness \cite{zhang2023provable}. Additionally, contrastive learning-based approaches have been proposed to refine unimodal embeddings by quantifying potential noise in unimodal data, thereby enhancing stability in multimodal fusion \cite{gao2024embracing}. Furthermore, some works introduce modality contribution evaluation, leveraging sample-wise reweighting via game theory \cite{wei2024enhancing} or modeling modality combinations as probabilistic distributions rather than fixed latent representations \cite{wei2024robust}.

While joint training enhances multimodal performance, it often limits unimodal capabilities, making models less robust in missing-modality scenarios. Recognizing that unimodal competence is the foundation for robust multimodal learning, \textbf{alternating training} frameworks have been proposed. Instead of joint optimization, these methods train each modality separately in a stepwise manner. \cite{zhang2024multimodal} introduce the Multimodal Learning with Alternating Unimodal Adaptation (MLA) framework, which minimizes cross-modal interference by decoupling unimodal and multimodal learning phases. \cite{hua2024reconboost} explore a modality-alternating training paradigm that mitigates modality competition by sequentially updating unimodal learners with a reconcilement regularization.

In this work, we tackle the challenges of modality imbalance and under-optimization in multimodal learning by introducing a dynamic weak-to-strong training strategy. By prioritizing weaker modalities and progressively enhancing cross-modal alignment, our method ensures balanced optimization and more effective modality interaction than conventional training paradigms.

\section{Preliminaries}
\noindent In this section, we formalize the multimodal learning setting by defining the input-output structure, training objectives, and optimization schemes. We differentiate between joint training and alternating training strategies, clarifying their mathematical formulations and implications. Beyond setup and comparison, we also introduce the \textbf{EDM}—a new modality-level diagnostic tool that quantifies alignment and capacity imbalance across modalities based on their contributions. This unified formulation lays the groundwork for our method and facilitates theoretical analysis in subsequent sections.

\subsection{Multimodal Learning Setup} 
\label{sec:mls}

\noindent We consider a multimodal learning framework with $n$ modalities, where each input sample consists of a set of modality-specific data representations:
\begin{equation}
\mathcal{M} = \{\mathbf{M}^{(1)}, \mathbf{M}^{(2)}, ..., \mathbf{M}^{(n)}\}, \quad \mathbf{Y} \sim \mathcal{D}
\end{equation}
where $\mathbf{M}^{(i)}=[\mathbf{M}^{(i)}_1,\mathbf{M}^{(i)}_2, ...,\mathbf{M}^{(i)}_m]\in \mathcal{}{R}^{ d_i\times m}$ represents the input from modality $i$, and $\mathbf{Y}$ is the ground truth label sampled from the data distribution $\mathcal{D}$. Each modality is processed through an encoder $f_i: \mathbf{M}^{(i)} \to \mathbf{X}^{(i)}$, mapping the raw input into a feature representation:
\begin{equation}
\mathbf{X}^{(i)} = f_i(\mathbf{M}^{(i)}; \theta_i)
\end{equation}
where $\theta_i$ are the parameters of the modality-specific encoder. The final multimodal representation $\mathbf{Z}$ is obtained by fusing these unimodal embeddings:
\begin{equation}
\mathbf{Z} = \text{Fusion}(\mathbf{X}^{(1)}, \mathbf{X}^{(2)}, ..., \mathbf{X}^{(n)})
\end{equation}
A classifier $g$ then maps the fused representation to the output space:
\begin{equation}
\mathbf{\widehat{Y}} = g(\mathbf{Z}; \zeta)
\end{equation}
where $\zeta$ are the classifier parameters. The model is trained to minimize the loss function $\ell(\mathbf{\widehat{Y}}, \mathbf{Y})$, typically cross-entropy for classification tasks.

\begin{figure*}[!t]
  \centering
   \includegraphics[width=0.8\linewidth]{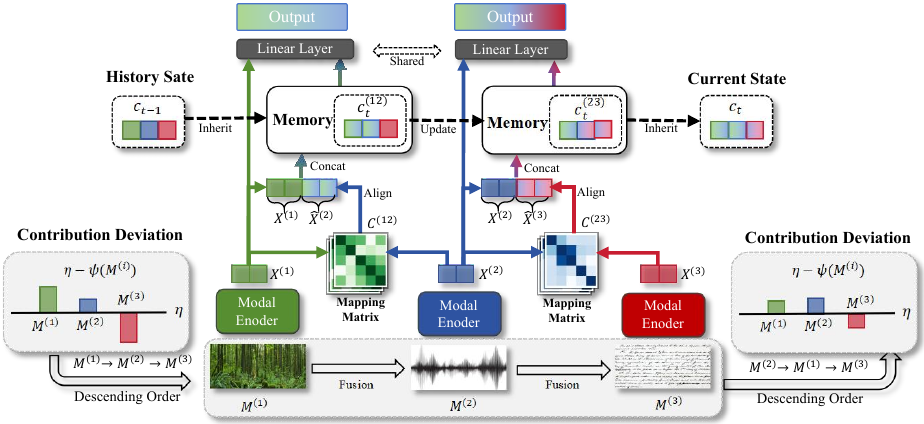}

   \caption{\textbf{Architecture of the proposed weak-to-strong alternating training framework.} Modalities are sequentially optimized based on contribution deviation, with memory carried across steps to enable progressive fusion and correction.}
   \label{fig:onecol}
\end{figure*}

\subsection{Joint vs. Alternating Training}  

\noindent \textbf{Joint Training} simultaneously optimizes all modality encoders and the classifier by minimizing the overall multimodal loss:
\begin{equation}
\mathcal{L}_{\text{joint}} = \mathbb{E}_{(\mathcal{M}, \mathbf{Y}) \sim \mathcal{D}} \left[ \ell(g(\text{Fusion}(\{f_i(\mathbf{M}^{(i)})\}_{i=1}^{n})), \mathbf{Y}) \right]
\end{equation}
where the gradients for all modality encoders \( \theta_1, \theta_2, ..., \theta_n \) are updated simultaneously in each training step. While this approach effectively captures multimodal interactions, it may introduce modality bias, where stronger modalities dominate learning and weaker ones contribute less.

\noindent \textbf{Alternating Training}, by contrast, optimizes modalities sequentially, updating one modality at a time while keeping others fixed. At each training epoch, a specific modality $\mathbf{M}^{(i)}$ is selected, and the model is updated using only that modality’s data:
\begin{equation}
    \mathcal{L}_{\text{alt}} = \sum_{i=1}^{n} \mathbb{E}_{(\mathcal{M}, \mathbf{Y}) \sim \mathcal{D}} \left[ \ell(g(\text{Fusion}(f_i(\mathbf{M}^{(i)}))), \mathbf{Y}) \right]
\end{equation}
This process reduces direct multimodal interference and ensures the enhancement of unimodal capabilities, but overlooks cross-modal correlations and structural consistency.

\subsection{Equilibrium Deviation Metric}

\noindent The EDM measures the extent to which modality contributions deviate from the ideal balanced state. Formally, let $\psi(\mathbf{M}^{(i)})$ denote the absolute contribution of modality $\mathbf{M}^{(i)}$ to the model’s performance. Then, the EDM is defined as:
\begin{equation}
\text{EDM}(\psi(\mathbf{M})) = \sum_{i=1}^{n} \left| \eta - \psi(\mathbf{M}^{(i)}) \right|
\end{equation}
where $\eta$ denotes the ideal contribution value for perfect equilibrium. This metric reflects the deviation from uniform contribution, with a smaller value indicating more balanced learning across modalities. In the ideal case where all modalities contribute equally, $\text{EDM} = 0$.

In our implementation, we compute $\psi(\mathbf{M}^{(i)})$ using a cooperative game-theoretic approach \cite{shapley1953value} based on marginal performance gains from different modality subsets:
\begin{equation}
\footnotesize
\psi(\mathbf{M}^{(i)}) = \sum_{\mathbf{S} \subseteq \mathcal{M} \setminus \{\mathbf{M}^{(i)}\}} 
\frac{|\mathbf{S}|! (n - |\mathbf{S}| - 1)!}{n!}\times \\ \left[ \mathcal{P}(\mathbf{S} \cup \{\mathbf{M}^{(i)}\}) - \mathcal{P}(\mathbf{S}) \right]
\end{equation}

Here, $\mathcal{P}(\mathbf{S})$ denotes the model's performance when only modalities in subset $\mathbf{S}$ are used for prediction. This ensures that $\psi(\mathbf{M}^{(i)})$ reflects each modality's marginal utility across all interaction contexts. We set $\eta = 100\%$ to reflect a balanced multimodal configuration.

\section{Methodology}
\noindent In this section, we present an \textit{EDM}-driven alternating training framework designed to dynamically balance modality contributions and promote effective multimodal integration. As illustrated in figure \ref{fig:onecol}, our method consists of three core components: (1) \textbf{Alternating Training Strategy}, which leverages the proposed EDM to adaptively determine the training order and balance modality-specific learning; (2) \textbf{Cross-Modal Alignment}, which facilitates deep feature interaction through a modality mapping layer and a modality-aware memory unit; and (3) \textbf{Memory-Guided Inference}, which enables efficient and consistent inference by allowing memory states to be progressively inherited across modalities. 

\label{sec:approach}

\subsection{Alternating Training Strategy}

\noindent Motivated by the bottleneck phenomenon in multimodal learning, in which the weakest modality constrains joint optimization, we draw inspiration from the “bucket effect” to develop a modality-wise alternating training strategy. Unlike conventional joint training that updates all modalities simultaneously, our approach updates them sequentially, allowing finer control over optimization dynamics. This design is grounded in both theoretical and empirical evidence, showing that the update order in alternating learning plays a decisive role. In particular, an EDM-guided weak-to-strong sequence proves most effective, yielding the tightest convergence bound and driving the model toward modality equilibrium. To formalize this, we introduce the following theorem:

\begin{theorem}
Weak-to-Strong ordering improves training under modality imbalance with memory transfer.

Consider a fusion objective $\mathcal{L}_{\text{fusion}}(\theta) = \alpha_1 \mathcal{L}_1(\theta, \mathbf{H}^{(1)}) + \alpha_2 \mathcal{L}_2(\theta, \mathbf{H}^{(2)})$, where $\alpha_1 \ll \alpha_2$ captures modality imbalance (e.g., text $<$ image), and $\mathbf{H}^{(1)} \rightarrow \mathbf{H}^{(2)}$ denotes knowledge transfer from weak to strong modality via memory.

If the stronger modality is better at absorbing errors (i.e., $\delta(\mathbf{H}^{(2)}) < \delta(\mathbf{H}^{(1)})$), then we have:
\begin{equation}
\mathcal{L}_{\text{fusion}}^{w \rightarrow s} < \mathcal{L}_{\text{fusion}}^{s \rightarrow w}
\end{equation}

\textbf{Proof sketch:} See Appendix A.
\end{theorem}

Following the theoretical insights discussed above, our alternating training strategy is implemented as follows: 

\noindent  \textbf{Modality Contribution Estimation and Scheduling.}
At the end of each training epoch, we estimate the contribution of each modality $\mathbf{M}^{(i)}$ by computing the deviation $\eta - \psi(\mathbf{M}^{(i)})$. This deviation reflects how much each modality lags behind the desired contribution. Modalities are then ranked in descending order of $\eta - \psi(\mathbf{M}^{(i)})$, forming the updated training schedule for the next epoch:
\begin{equation}
    \mathbf{M}_{\text{priority}} = \operatorname{arg\,sort}_{\mathbf{M}^{(i)} \in \mathcal{M}} (\eta -\psi(\mathbf{M}^{(i)}))
\end{equation}

\noindent \textbf{Sequential Gradient Update.} 
Once the ordered modality list $\mathbf{M}_{\text{priority}}$ is determined, we apply gradient-based optimization in a pairwise sequential manner. Specifically, we group the sorted modalities into adjacent pairs. For each pair $(\mathbf{M}^{(i)}, \mathbf{M}^{(j)})$, we first update the model using $\mathbf{M}^{(i)}$, and then use the past state $\mathbf{H}^{(i)}$ to guide the update of $\mathbf{M}^{(j)}$:
\begin{equation}
\begin{aligned}
\theta \leftarrow \theta - \nabla \mathcal{L}(\mathbf{M}^{(i)}; \theta), \quad 
\mathbf{H}^{(i)} \rightarrow \mathbf{H}^{(j)}, \quad \\
\theta \leftarrow \theta - \nabla \mathcal{L}(\mathbf{M}^{(j)}, \mathbf{H}^{(j)}; \theta)
\end{aligned}
\end{equation}

The memory is then passed to the next modality pair to ensure continuity across training. This design allows later modalities to benefit from earlier ones, aligning with the weak-to-strong optimization principle.

\subsection{Cross-Modal Alignment}
\label{sec:cma}
\noindent To facilitate information transmission across modalities during alternating training, we propose a cross-modal alignment module that integrates weak-to-strong feature fusion and long-term memory propagation. Unlike prior approaches such as MLA \cite{zhang2024multimodal} that suppress gradient interactions via orthogonalization, we explicitly encourage useful information flow between modalities to improve cross-modal consistency and overall integration.

\noindent \textbf{Feature and Sample-Level Alignment.} Given a training schedule that orders modalities from weak to strong, we align features between modality pairs to enable effective fusion. For any two modalities $\mathbf{M}^{(i)}$ and $\mathbf{M}^{(j)}$ ($i < j$), we compute a sample-wise correlation matrix:
\begin{equation}
    \mathbf{C}^{(ij)}_{rq} = \frac{<{\mathbf{X}_r^{(i)}},\mathbf{X}_q^{(j)}>}{\|\mathbf{X}_r^{(i)}\| \|\mathbf{X}_q^{(j)}\|}
\end{equation}
where $\mathbf{X}^{(i)}$ and $\mathbf{X}^{(j)}$ denote the feature matrices for modalities $\mathbf{M}^{(i)}$ and $\mathbf{M}^{(j)}$ within a batch. We retain only high-confidence relations with a threshold $\tau$:
\begin{equation}
    \mathbf{C}^{(ij)} = \mathbf{C}^{(ij)} \odot \mathbf{1}(\mathbf{C}^{(ij)} > \tau), \quad \mathbf{C}^{(ij)}_{qq} = 1, \forall q
\end{equation}
where $\odot$ denotes element-wise multiplication, and the diagonal is set to 1 to preserve original features during fusion. The aligned representation from the stronger modality $\widehat{\mathbf{X}}^{(j)}$ is then computed as:
\begin{equation}
    \widehat{\mathbf{X}}^{(j)} = \mathbf{X}^{(j)}\mathbf{C}^{(ij)} 
\end{equation}

\noindent \textbf{Modality-Aware Memory Unit.}
To enable progressive fusion across both modality pairs and training epochs, we introduce a long-term memory module that maintains and updates a memory state across modalities and across samples over time. Inspired by sequential models such as LSTM \cite{schmidhuber1997long}, this module allows each weaker modality to selectively pass relevant information to a stronger aligned modality, while preserving historical context from prior interactions. Importantly, the memory not only flows between modality pairs within the same epoch but is also propagated across training epochs to support long-range consistency.

We maintain a memory state $\mathbf{c}_{t}^{(ij)}$, which aggregates context-aware information from modality interactions over sequential training steps. This state is propagated from the previous aligned modality pair at step $t-1$ or from the last epoch (if $t = 0$). The update process is detailed as follows:

\noindent1) Directional feature fusion from modality $\mathbf{M}^{(i)}$ and its aligned counterpart $\mathbf{M}^{(j)}$:
\begin{equation}
\mathbf{z}^{(ij)} = \operatorname{concat}\left(\mathbf{H}^{(i)}, \widehat{\mathbf{X}}^{(j)}\right)
\end{equation}
Here, For the first modality $\mathbf{M}^{(i)}$ in the sequence, where no memory is available yet, we set $\mathbf{H}^{(i)} = \mathbf{X}^{(i)}$.

\noindent2) Gate computations to regulate information flow:
\begin{equation}
\begin{aligned}
\mathbf{f}^{(ij)} &= \sigma\left(\mathbf{W}_f \mathbf{z}^{(ij)}\right), \\
\mathbf{i}^{(ij)} &= \sigma\left(\mathbf{W}_i \mathbf{z}^{(ij)}\right), \\
\mathbf{o}^{(ij)} &= \sigma\left(\mathbf{W}_o \mathbf{z}^{(ij)}\right)
\end{aligned}
\end{equation}
where $\mathbf{f}^{(i)}$ retains modality-specific memory, $\mathbf{i}^{(i)}$ controls cross-modal updates, and $\mathbf{o}^{(i)}$ regulates fusion output.

\noindent3) Candidate state derived from cross-modal interactions:
\begin{equation}
\widetilde{\mathbf{g}}^{(ij)} = \tanh\left(\mathbf{W}_g \mathbf{z}^{(ij)}\right)
\end{equation}
4) Context-aware memory update:
\begin{equation}
\mathbf{c}_{t}^{(ij)} = \mathbf{f}^{(ij)} \odot \mathbf{c}_{t-1}^{(ij)} 
+ \mathbf{i}^{(ij)} \odot \widetilde{\mathbf{g}}^{(ij)}
\end{equation}
5) Fused representation for modality $\mathbf{M}^{(j)}$ at time $t$:
\begin{equation}
\mathbf{H}^{(j)} = \mathbf{o}^{(ij)} \odot \tanh\left(\mathbf{c}_{t}^{(ij)}\right)
\end{equation}

\subsection{Memory-Guided Inference}

\noindent During inference, the absence of supervision makes it infeasible to dynamically estimate modality contributions. To address this, we propose a memory-inherited weak-to-strong inference strategy, which reuses the modality update order derived from the last training epoch, assumed to reflect stable modality imbalance patterns (discussed in supplemental materials).

Specifically, given input modalities ${\mathbf{M}^{(1)}, \mathbf{M}^{(2)}, \mathbf{M}^{(3)}}$, we follow the weak-to-strong update schedule obtained at the end of training (e.g., ${1 \rightarrow 2 \rightarrow 3}$). Each modality is processed sequentially using the memory-based fusion module introduced above, where the output of each step is used to inform the next through the long-term memory state $\mathbf{c}_t$.

At the first step, the weakest modality $\mathbf{M}^{(1)}$ is directly encoded to obtain its initial representation $\mathbf{H}^{(1)}$. In subsequent steps, each stronger modality is fused with the previous representation via the memory module, progressively accumulating multimodal knowledge. The final representation $\mathbf{H}^{(3)}$—obtained from the strongest modality—serves as the unified multimodal embedding, which is then projected to the output space for prediction. The complete inference procedure is summarized in Algorithm~\ref{alg:inference}.

\begin{algorithm}
\caption{Memory-Inherited Weak-to-Strong Inference}
\label{alg:inference}
\textbf{Input:} Modal inputs $\{\mathbf{M}^{(1)}, \mathbf{M}^{(2)}, \dots, \mathbf{M}^{(n)}\}$, modality priority order $\mathbf{M}_{\text{priority}}$ \\
\textbf{Output:} Prediction $\mathbf{\widehat{Y}}$
\begin{algorithmic}[1]
\STATE Initialize memory: $\mathbf{c}_{0} \gets \mathbf{0}$
\FOR{$i = 1$ to $n$}
    \STATE Let $k \gets \mathbf{M}_{\text{priority}}[i]$ \COMMENT{Get current modality index}
    \STATE Extract features: $\mathbf{X}^{(k)} \gets f_k(\mathbf{M}^{(k)}; \theta_k)$
    \STATE Set $\mathbf{H}^{(k)} \gets \mathbf{X}^{(k)}$
    \STATE Let $p \gets \mathbf{M}_{\text{priority}}[i-1]$
    \STATE Fuse: $(\mathbf{H}^{(k)}, \mathbf{c}_{i}) \gets \text{Memory}(\mathbf{H}^{(p)}, \mathbf{X}^{(k)}, \mathbf{c}_{i-1})$
\ENDFOR
\STATE Compute prediction: $\mathbf{\widehat{Y}} \gets g(\mathbf{H}^{(k)}; \zeta)$
\RETURN $\mathbf{\widehat{Y}}$
\end{algorithmic}
\end{algorithm}

\section{Experiments}
\label{sec:experiment}

\begin{table*}
  \centering
  \caption{\textbf{Comparison of performance across four datasets.} \textbf{Bolded} numbers indicate the best results, while \underline{underlined} numbers represent the second-best performance. $\dag$ indicates the subset.}
  \small
  \setlength{\tabcolsep}{1mm} 
  \scalebox{0.85}{
  \begin{tabular}{@{}c|lc|cc|cc|cc|cc|cc|cc|cc|cc|cc}
    \toprule
   Type & Data & \diagbox[width=5em]{$Acc$}{$EDM$}& \multicolumn{2}{c|}{Sum}  &\multicolumn{2}{c|}{Concat}  & \multicolumn{2}{c|}{OGM}  & \multicolumn{2}{c|}{OGM-GE}  & \multicolumn{2}{c|}{LFM}  & \multicolumn{2}{c|}{CRMT} & \multicolumn{2}{c|}{Late Fusion} & \multicolumn{2}{c|}{MLA} & \multicolumn{2}{c}{Ours} \\
    \midrule
  \multirow{9}{*}{{\color{red}{T}}-{\color{gre}{V}}} & \multirow{2}{*}{MVSA}      & Text  & \color{red}{75.34}& \color{red}{3.18} &\color{red}{75.72}& \color{red}{0.96}& \color{red}{75.53} & \color{red}{2.70}&\color{red}{75.14}& \color{red}{14.92}&\color{red}{74.37}& \color{red}{1.51}&\color{red}{52.41}&\color{red}{44.82}&\color{red}{75.34}& \color{red}{1.54}&\color{red}{74.76}&\color{red}{2.22} &\color{red}{72.45}&{\color{red}{0.45}}\\
                            &  &Visual & \color{gre}{30.25}&\color{gre}{31.48}& \color{gre}{52.02}& \color{gre}{31.70} &\color{gre}{34.68}&\color{gre}{35.88} &\color{gre}{31.60}& \color{gre}{39.49}&\color{gre}{57.23}& \color{gre}{2.93}&\color{gre}{52.02}& \color{gre}{44.82}&\color{gre}{57.42}&\color{gre}{12.22}&\color{gre}{57.42}& \color{gre}{2.48}&\color{gre}{54.14}&{\color{gre}{1.74}}\\
                            & (M3AE) &\textbf{Multi} & 76.30& {34.66}& 75.92&{32.67}&  75.72&{38.58}& 75.72&{54.41}& 75.53&\underline{4.44}& 54.53&{89.64}& 76.30& {28.68}&\underline{77.46}&{4.69}& \textbf{78.42}& \textbf{2.19} \\
    \cmidrule{2-21}
    & \multirow{2}{*}{MVSA}&  Text  & \color{red}{76.30}& \color{red}{3.15} &\color{red}{74.76}& \color{red}{0.42}& \color{red}{73.80} & \color{red}{0.19}&\color{red}{74.57}& \color{red}{0.13}&\color{red}{75.72}& \color{red}{0.06}&\color{red}{72.06}&\color{red}{1.54}&\color{red}{75.14}& \color{red}{34.05}&\color{red}{73.60}&\color{red}{0.16} &\color{red}{75.53}&{\color{red}{0.06}}\\
                            &  &Visual & \color{gre}{62.04}&\color{gre}{0.19}& \color{gre}{63.39}& \color{gre}{0.93} &\color{gre}{63.58}&\color{gre}{2.64} &\color{gre}{65.90}& \color{gre}{0.38}&\color{gre}{68.02}& \color{gre}{0.19}&\color{gre}{72.25}& \color{gre}{1.29}&\color{gre}{69.75}&\color{gre}{65.50}&\color{gre}{69.75}& \color{gre}{1.32}&\color{gre}{70.52}&{\color{gre}{0.19}}\\
                            & (Ola-7B) &\textbf{Multi} & 78.81& {3.34}&79.38 &{1.35}& 79.96 &{2.83}& \underline{80.54}&\underline{0.51}& 79.77&\textbf{0.25}& 72.64&{2.83}& 79.58& {99.55}&{79.96}&{1.48}& \textbf{80.92}& \textbf{0.25} \\
    \cmidrule{2-21}
    & \multirow{2}{*}{MVSA}&  Text  & \color{red}{32.37}& \color{red}{25.18} &\color{red}{34.68}& \color{red}{13.63}& \color{red}{53.37} & \color{red}{28.84}&\color{red}{53.56}& \color{red}{30.29}&\color{red}{47.59}& \color{red}{12.22}&\color{red}{67.05}&\color{red}{24.37}&\color{red}{48.75}& \color{red}{62.22}&\color{red}{51.83}&\color{red}{15.88} &\color{red}{52.99}&\color{red}{15.85}\\
                            &  &Visual & \color{gre}{64.35}&\color{gre}{5.11}& \color{gre}{76.30}& \color{gre}{2.32} &\color{gre}{77.07}&\color{gre}{4.21} &\color{gre}{73.80}& \color{gre}{6.37}&\color{gre}{77.26}& \color{gre}{5.92}&\color{gre}{69.36}& \color{gre}{24.89}&\color{gre}{77.26}&\color{gre}{31.16}&\color{gre}{79.00}& \color{gre}{7.27}&\color{gre}{74.76}&{\color{gre}{0.16}}\\
                            & (LLaVA-7B) &\textbf{Multi} & 78.42& 30.29& 77.84& \textbf{15.95}& 78.61 & 33.05& \underline{79.19} & 36.65& 76.69& {18.14}& 73.03& 49.26& 78.42& 93.38& 79.00& 23.15& \textbf{79.96}& \underline{16.01} \\
    \midrule
   \multirow{9}{*}{{\color{blue}{A}}-{\color{gre}{V}}} &  \multirow{2}{*}{CREMA-D}  & Audio & \color{blue}{57.66}&\color{blue}{5.98}& \color{blue}{56.45}&\color{blue}{2.53}&  \color{blue}{61.96}&\color{blue}{0.09}& \color{blue}{54.84}&\color{blue}{1.17}& \color{blue}{58.47}&\color{blue}{1.05}& \color{blue}{56.99}& \color{blue}{6.82}&\color{blue}{58.87}&\color{blue}{71.80}&\color{blue}{58.74}&\color{blue}{2.34}& \color{blue}{55.78}&\color{blue}{2.19}\\
                            &  &Visual & \color{gre}{35.62}&\color{gre}{26.61}& \color{gre}{39.38}&\color{gre}{35.76}&  \color{gre}{39.25}&\color{gre}{36.55}& \color{gre}{38.98}&\color{gre}{33.51}& \color{gre}{72.58}&\color{gre}{6.75}& \color{gre}{57.66}&\color{gre}{6.90}&\color{gre}{71.24}&\color{gre}{72.48}&\color{gre}{73.12}&\color{gre}{4.46}&\color{gre}{71.37}&\color{gre}{0.59}\\
                            & (ResNet) &\textbf{Multi} & 70.30&{32.59}& 70.30&{38.29}&  72.98&{36.64}& 70.30&{34.68}& \underline{81.59}&{7.79}& 71.10&{13.72}&78.63&144.28&81.05&\underline{6.79}& \textbf{84.95}&\textbf{2.78}\\
    \cmidrule{2-21}
    &  \multirow{2}{*}{CREMA-D$^\dag$}  & Audio & \color{blue}{41.13}&\color{blue}{0.40}& \color{blue}{45.16}&\color{blue}{1.48}&  \color{blue}{50.81}&\color{blue}{0.04}& \color{blue}{48.39}&\color{blue}{6.00}& \color{blue}{52.42}&\color{blue}{1.57}& \color{blue}{25.00}& \color{blue}{60.35}&\color{blue}{37.90}&\color{blue}{83.02}&\color{blue}{45.16}&\color{blue}{0.36}& \color{blue}{49.19}&\color{blue}{3.76}\\
                            &  &Visual & \color{gre}{17.74}&\color{gre}{23.52}& \color{gre}{21.77}&\color{gre}{28.00}&  \color{gre}{24.19}&\color{gre}{38.93}& \color{gre}{20.16}&\color{gre}{34.14}& \color{gre}{22.58}&\color{gre}{29.26}& \color{gre}{20.97}&\color{gre}{59.54}&\color{gre}{23.39}&\color{gre}{82.93}&\color{gre}{23.39}&\color{gre}{35.30}&\color{gre}{23.39}&\color{gre}{9.32}\\
                            & (Ola-7B) &\textbf{Multi} & 55.65&\underline{23.92}& 57.26&{29.48}& \underline{58.06} &{38.97}& 55.65 &{40.14}& {40.32}&{30.82}& 27.42&{119.89}&56.45&165.95&{52.42}&{35.66}& \textbf{58.87}&\textbf{13.08}\\
                            
    \cmidrule{2-21}
    & \multirow{2}{*}{Kinetics-400} &      Audio & \color{blue}{49.01}&\color{blue}{0.97}& \color{blue}{51.59}&\color{blue}{0.21}& \color{blue}{50.64}&\color{blue}{0.64}& \color{blue}{47.73}&\color{blue}{1.58}& \color{blue}{57.07}&\color{blue}{4.21}& \color{blue}{57.16}&\color{blue}{8.94}& \color{blue}{54.16}&\color{blue}{94.34}&\color{blue}{52.70}&\color{blue}{5.56}&\color{blue}{53.21}&\color{blue}{0.20}\\
                           &  & Visual &\color{gre}{33.08} &\color{gre}{29.81}& \color{gre}{36.08} &\color{gre}{31.35}& \color{gre}{41.05}&\color{gre}{31.23}& \color{gre}{38.99}&\color{gre}{30.79}& \color{gre}{63.67}&\color{gre}{4.89}&\color{gre}{61.01} &\color{gre}{10.86}& \color{gre}{59.04}&\color{gre}{94.85}&\color{gre}{58.87}&\color{gre}{5.95} &\color{gre}{63.67}&\color{gre}{0.11}\\
                           & (ResNet) & \textbf{Multi} &66.07 &{30.78}&67.61 &{31.56}& 68.89&{31.87}& 65.98&{32.37}& 75.15&\underline{9.10}& \underline{75.84} &{21.08}& 71.72&{189.19}&72.66&{11.51}&\textbf{79.35}&\textbf{0.31}\\
    \midrule
   \multirow{4}{*}{{\color{blue}{A}}-{\color{red}{T}}-{\color{gre}{V}}} & \multirow{2}{*}{IEMOCAP}  & Audio &\color{blue}{46.43} &\color{blue}{3.49}& \color{blue}{44.81}&\color{blue}{2.17}& \color{blue}{46.61}&\color{blue}{4.46}& \color{blue}{48.60}&\color{blue}{2.96}& \color{blue}{51.22}&\color{blue}{1.36}& \color{blue}{55.65}&\color{blue}{0.33}&\color{blue}{53.93}&\color{blue}{0.43}& \color{blue}{52.94}&\color{blue}{0.00}&\color{blue}{53.48}&\color{blue}{5.09}\\
                           &  & Text & \color{red}{61.61}&\color{red}{14.40}& \color{red}{63.23}&\color{red}{16.56}&\color{red}{62.96}&\color{red}{14.15}&\color{red}{64.14}&\color{red}{14.49}& \color{red}{67.75}&\color{red}{5.21}& \color{red}{56.19}&\color{red}{0.43}&\color{red}{66.12}&\color{red}{11.49}& \color{red}{66.12}&\color{red}{11.63}&\color{red}{60.16}&\color{red}{1.79}\\
                          & \multirow{2}{*}{(M3AE)} & Visual & \color{gre}{26.83}&\color{gre}{14.89}& \color{gre}{28.36}&\color{gre}{14.39}& \color{gre}{30.71}&\color{gre}{14.39}& \color{gre}{22.67}&\color{gre}{14.06}& \color{gre}{30.44}&\color{gre}{26.90}& \color{gre}{30.44}&\color{gre}{30.07}&\color{gre}{30.71}&\color{gre}{21.09}& \color{gre}{30.44}&\color{gre}{20.40}&\color{gre}{30.71}&\color{gre}{11.71}\\
                           &  & \textbf{Multi} & 73.26&{32.78}& \underline{74.71}& \text{33.12}& 74.62&{33.00}& 73.98&{31.51}& 71.27&33.47& 55.92&\underline{30.83}&71.45&33.01& 72.36&32.04& \textbf{75.07}&\textbf{18.60}\\
    \bottomrule
  \end{tabular}}
  \label{tab:tab1}
\end{table*}

\subsection{Experimental Setup}

\noindent \textbf{Datasets.} We evaluate our approach on four widely used multimodal datasets covering diverse modality combinations and tasks. \textbf{MVSA-Single} \cite{niu2016sentiment}  is a text-image dataset, focusing on sentiment analysis. \textbf{CREMA-D} \cite{cao2014crema} and \textbf{Kinetics-400} \cite{arandjelovic2017look} are speech-image datasets, used for emotion recognition and action recognition based on speech and visual cues. Finally, \textbf{IEMOCAP} \cite{busso2008iemocap} is a tri-modal dataset integrating speech, text, and facial pictures.

\noindent\textbf{Baselines.} We compare our approach against two major categories of multimodal learning frameworks: joint training and alternating training. For joint training, we include simple methods such as \textbf{Sum} and \textbf{Concat}, as well as recent SOTA methods like \textbf{OGM/OGM-GE} \cite{peng2022balanced}, \textbf{LFM} \cite{yang2025facilitating}, and \textbf{CRMT} \cite{yang2024quantifying}, which enhance multimodal learning through gradient modulation and feature refinement. For alternating training, we consider \textbf{Late Fusion} \cite{gunes2005affect}, where modalities are trained separately and their predictions are combined at inference, and \textbf{MLA} \cite{zhang2024multimodal}, the most advanced alternating training framework that minimizes cross-modal interference via gradient decorrelation.

\noindent\textbf{Traditional Encoders Setting.} Our framework is instantiated with modality-specific encoders tailored to each dataset configuration. For MVSA-Single dataset, we employ M3AE-Large \cite{geng2022multimodal}; for CREMA-D and Kinetics-400 datasets, a ResNet-18-based encoder \cite{he2016deep} is used. In the tri-modal setting, we integrate M3AE-Base for text–image encoding and CAVMAE \cite{gongcontrastive} for speech. All models are trained using the AdamW optimizer with a learning rate of 1e-4.

\noindent\textbf{MLLM Setting.} To explore the effectiveness of our approach under more powerful backbone settings, we adopt Ola-7B \cite{liu2025ola} as a representative MLLM capable of processing text, image, and speech inputs. Unlike conventional encoders such as ResNet or M3AE, where each modality is encoded independently through shallow feature extractors, we leverage Ola-7B in a prompt-based fashion:

\begin{enumerate}
    \item For each modality input (text, image, or speech), we first extract modality-specific features using a pretrained encoder (e.g., CLIP-ViT-L/14 for images, Whisper-large-v3 for speech).
    \item These features are formatted into a prompt (default prompt used by Ola) and fed into the MLLM. The output corresponding to a designated token, denoted as \texttt{[CLASS]}, is used as the modality representation:
    \begin{equation}
    \mathbf{X}^{(i)} = \texttt{MLLM}(\texttt{Prompt}(\mathbf{M}^{(i)}))[\texttt{[CLASS]}]
    \end{equation}
    \item To adapt these representations for downstream alignment and classification, we apply a lightweight transformation consisting of a linear layer followed by ReLU activation:
    \begin{equation}
    \tilde{\mathbf{X}}^{(i)} = \texttt{ReLU}(\mathbf{W}_i \cdot \mathbf{X}^{(i)} + \mathbf{b}_i)
    \end{equation}
\end{enumerate}

This refined feature $\tilde{\mathbf{X}}^{(i)}$ is then passed to the memory and classification modules. All MLLM backbone parameters are frozen during training and inference step.

\noindent\textbf{Evaluation Metrics.} To comprehensively evaluate multimodal model performance, we report classification accuracy (\%) as the primary metric. Additionally, we introduce the EDM to quantify modality imbalance during integration. While accuracy reflects task effectiveness, EDM offers a complementary perspective on the robustness and balance of multimodal learning.

\subsection{Main Results}

\noindent \textbf{Results based on M3AE / ResNet.} Table \ref{tab:tab1} shows that our method consistently achieves SOTA fusion accuracy and the lowest EDM across all four datasets, with an average accuracy gain of 1.58. Notably, we observe +3.36 on CREMA-D and +3.51 on Kinetics-400, highlighting effective modality integration. These results support three key findings: (1) \textit{Unimodal strength alone is insufficient for effective fusion}. For example, Late Fusion methods achieve strong performance on individual modalities across datasets but yield significantly lower fusion accuracy, indicating poor cross-modal integration; (2) \textit{Lower EDM correlates with better fusion.} A clear trend emerges: models with lower equilibrium deviation scores, such as ours, LFM, and MLA, consistently achieve higher fusion accuracy, validating EDM as an indicator of modality integration quality; (3) \textit{Balance should be contribution-aware, not performance-equalized.} Methods like CRMT enforce equal unimodal accuracy but degrade fusion performance by ignoring inherent backbone disparities. Our strategy instead optimizes the contribution deviation of each modality, yielding better alignment and integration.

\noindent \textbf{Results based on Ola-7B.} Table \ref{tab:tab1} shows that with MLLMs as the backbone, our method consistently outperforms existing strategies across both MVSA-Single and CREMA-D$^\dag$. When training data is sufficient and the backbone model is strong, simple Sum and Concat strategies already achieve competitive results, with Sum reaching 78.81 and Concat 79.38 on MVSA-Single. In this setting, advanced multimodal learning strategies like LFM and MLA fail to show a clear advantage. However, our approach further refines modality balance, leading to an 80.92 fusion accuracy, surpassing all baselines. In contrast, when training data is scarce, existing methods degrade significantly, with MLA dropping to 52.42, performing no better than simple fusion methods. These results demonstrate that our approach remains effective even with limited data.

\noindent \textbf{Results based on LLaVA-7B.} To assess the generalizability of our method across different MLLM architectures, we further conduct experiments using LLaVA-7B as the backbone on the MVSA dataset. As shown in Table~\ref{tab:tab1}, our method consistently outperforms both joint training and existing alternating training baselines under the LLaVA-7B setting. Compared to Ola-7B, LLaVA-7B exhibits relatively weaker textual modeling capability, leading to a more pronounced imbalance between visual and textual representations. Despite this increased asymmetry, our approach effectively fuses multimodal information and achieves substantial performance improvements. These results indicate that the proposed strategy can robustly align heterogeneous modalities and remains effective even when the backbone model exhibits asymmetric modality strengths.

\begin{table}[t]
\centering
\caption{\textbf{Ablation study.} $\Delta$ Loss denotes the performance drop after removing a module. $\Delta$ Gain measures the improvement of multimodal fusion over the unimodal baseline.}
\small
\setlength{\tabcolsep}{1mm}
\begin{tabular}{l|cccc|c}
\toprule
\multicolumn{6}{c}{\textbf{Component Ablation (CREMA-D)}} \\
\midrule
\textbf{Model Variant} & Audio& Text& Visual& Multi & $\Delta$ Loss \\
\midrule
Full Model (Ours)& 55.78& - &\textbf{71.37}& \textbf{84.95} & - \\
w/o Alignment & 55.51&  -& 70.16&  \underline{84.41}& -0.54\\
w/o Memory Unit& 49.73& -& 64.25&  64.25& -20.70 \\
w/o Memory Inheritance& 54.57 & -&58.06& 78.36 & -6.59 \\
\midrule
\multicolumn{6}{c}{\textbf{Modality Ablation (IEMOCAP)}} \\
\midrule
\textbf{Modality} & Audio& Text& Visual& Multi & $\Delta$ Gain \\
\midrule
Audio Only & \textbf{55.47} &- &- & - & - \\
Text Only & - & \textbf{68.56}& -& - & - \\
Visual Only & - & -& \textbf{32.97}& - & - \\
Audio + Text & 52.48 & {64.23} & -& \underline{72.18} & +7.95 \\
Audio + Visual & 51.04 & -& \underline{30.71}& 55.92 & +4.88 \\
Text + Visual & - & \underline{65.76}& 30.44& 65.85 & +0.09 \\
All Modalities (Ours) & 53.48& 60.16& \underline{30.71}&\textbf{75.07} & +14.89 \\
\bottomrule
\end{tabular}

\label{tab:tab3}
\end{table}

\label{sec:ablation}
\subsection{Ablation Study}

\noindent\textbf{Component Ablation.} Table \ref{tab:tab3} (top) shows ablation results on the CREMA-D dataset, validating the effectiveness of each component. Removing the cross-modal mapping slightly reduces fusion accuracy from 84.95 to 84.41, indicating that structural misalignment impairs integration. Eliminating the memory module leads to a sharp drop in unimodal performance, where audio accuracy decreases from 55.78 to 49.73 and visual accuracy drops from 71.37 to 64.25. Furthermore, removing the memory inheritance mechanism causes a notable decrease of 6.59 in multimodal accuracy, highlighting its importance in propagating cross-modal knowledge across modality updates.

\noindent\textbf{Modality Ablation.} Table~\ref{tab:tab3} (bottom) illustrates the benefits of modality fusion. While unimodal models yield limited performance (Audio Only: 55.47, Text Only: 68.56), combining the two boosts accuracy to 72.18, indicating that even weaker modalities can enhance results when properly integrated. As more modalities are added, performance steadily improves: the full three-modality setup achieves 75.07, exceeding the best bimodal configuration by 2.89. These results validate the effectiveness of our memory-guided strategy in fully leveraging cross-modal complementarity.

\begin{figure}[!t]
    \centering
    \subfloat[\footnotesize Initial order comparison]{
        \centering
        \includegraphics[width=0.48\linewidth]{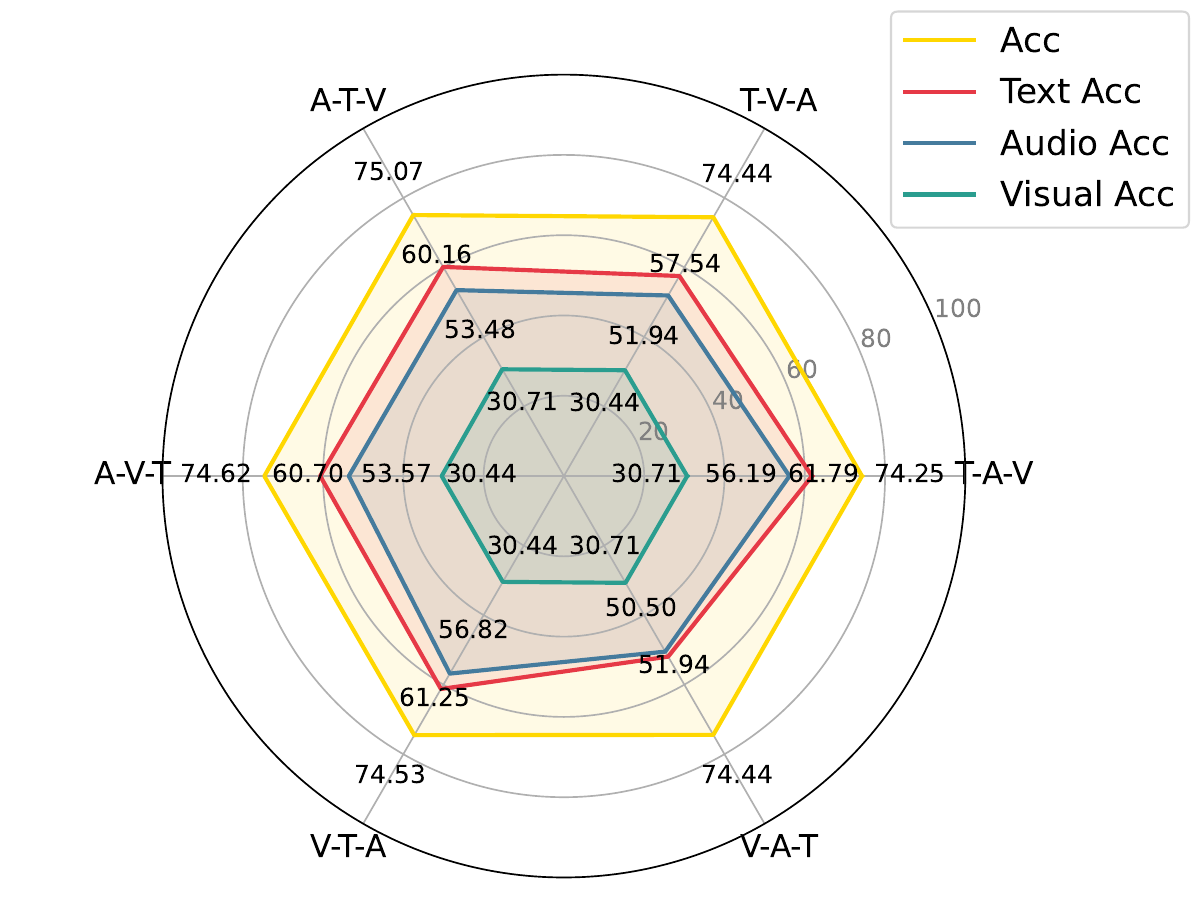}
        }
    \subfloat[\footnotesize Ordering method comparison]{
        \centering
        \includegraphics[width=0.48\linewidth]{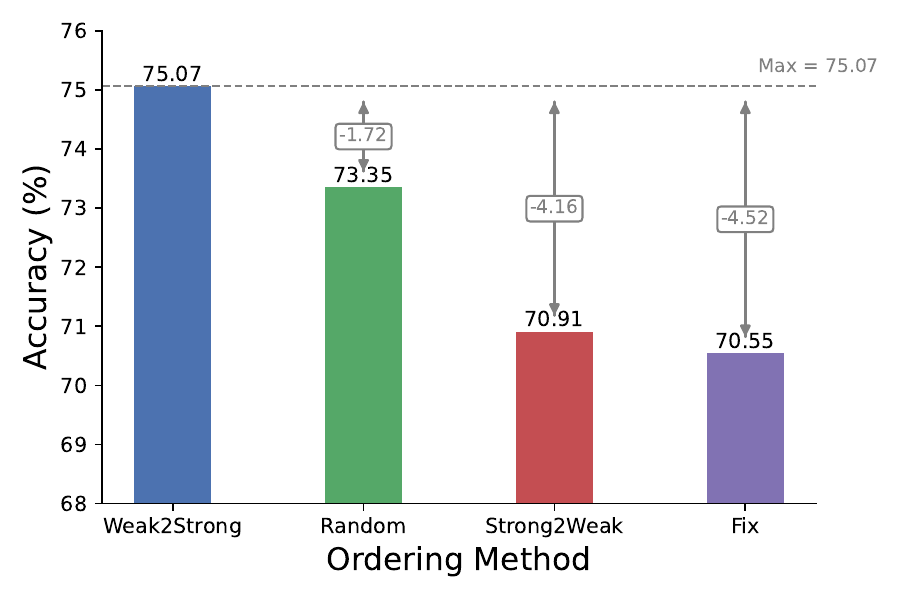}
        }
    \caption{\textbf{Order analysis.} Comparison of initial update sequences and order scheduling strategies.}
    \label{fig:radar}
\end{figure}

\subsection{Model Analysis}
\label{analysis}
\noindent\textbf{Modality Order Analysis.} We analyze the role of modality update order in our memory-guided alternating training framework from two perspectives: (a) the impact of initial ordering and (b) the influence of update strategy on final performance. As shown in figure~\ref{fig:radar}(a), different initial orders result in less than 1\% fluctuation in final multimodal accuracy, indicating that the model quickly adjusts its update schedule based on contribution deviation after the early training epochs. However, we observe that placing stronger modalities (e.g., Text or Audio) earlier accelerates memory initialization and leads to faster convergence. Figure~\ref{fig:radar}(b) compares different update strategies. The weak-to-strong ordering consistently outperforms others across benchmarks, while the strong-to-weak strategy even underperforms random order. These results confirm our hypothesis: during memory inheritance, strong modalities are more effective at absorbing accumulated errors than weak ones are at correcting them—an effect we term \textit{Error Absorption} (i.e., $\delta(\mathbf{H}^{(1)})$) versus \textit{Error Amplification} (i.e., $\delta(\mathbf{H}^{(2)})$).

\begin{figure}[!t]
    \centering
    \subfloat[\footnotesize M3AE encoder]{
        \centering
        \includegraphics[width=0.48\linewidth]{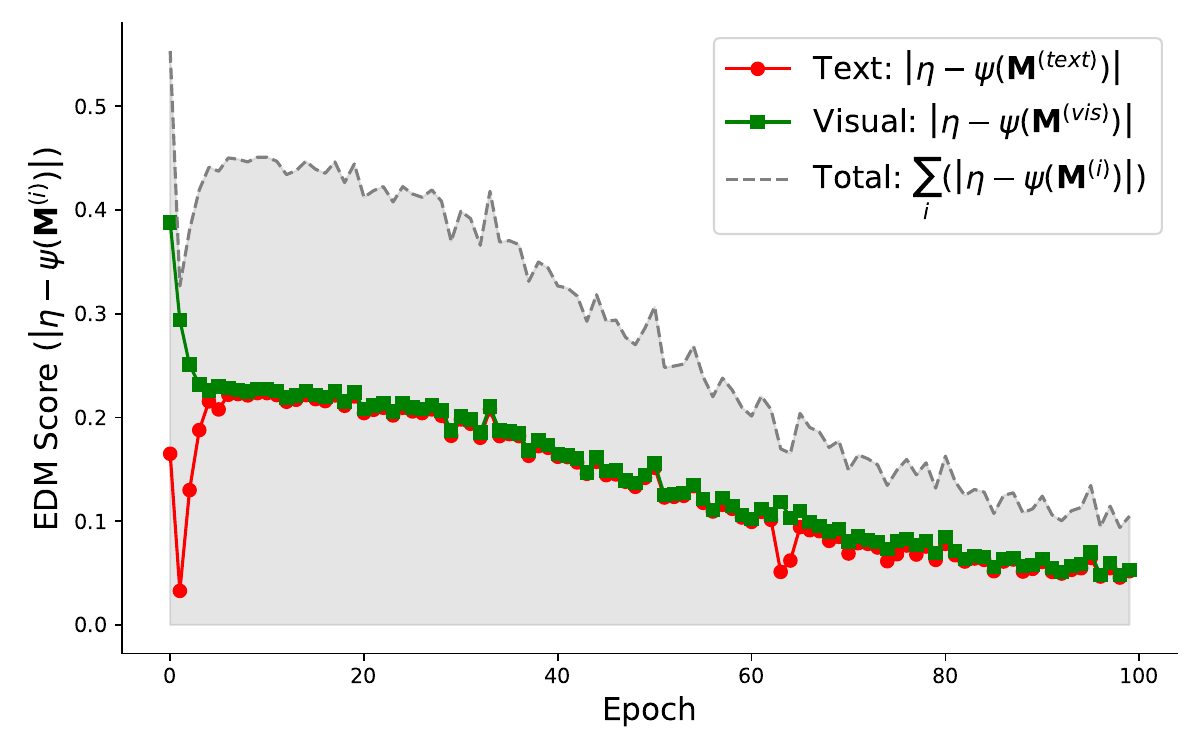}
        }
     \subfloat[\footnotesize Ola-7B encoder]{
        \centering
        \includegraphics[width=0.48\linewidth]{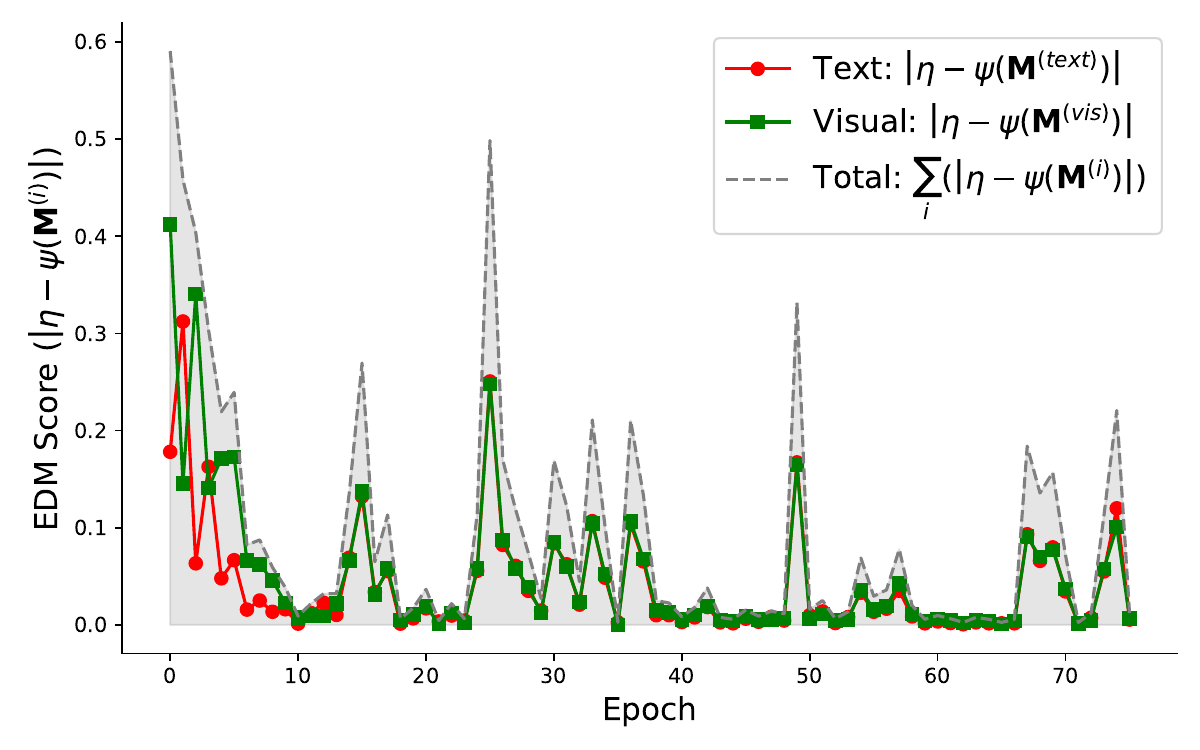}
        }
    \caption{\textbf{EDM score analysis.} Modality contribution deviation shifts under varying encoders.}
    \label{fig:modality_contribution}
\end{figure}

\noindent\textbf{EDM Score Analysis.}
Figure \ref{fig:modality_contribution} illustrates the evolution of EDM scores on MVSA-Single, showing that despite initial imbalances due to model-specific strengths, our training strategy consistently drives the model toward equilibrium. When the initial gap is large, the framework redistributes contributions by slightly reducing the dominant modality, whereas when both modalities are strong, stepwise optimization enhances both in a mutually reinforcing manner. 

\begin{figure}[!t]
    \centering
    \includegraphics[width=0.95\linewidth]{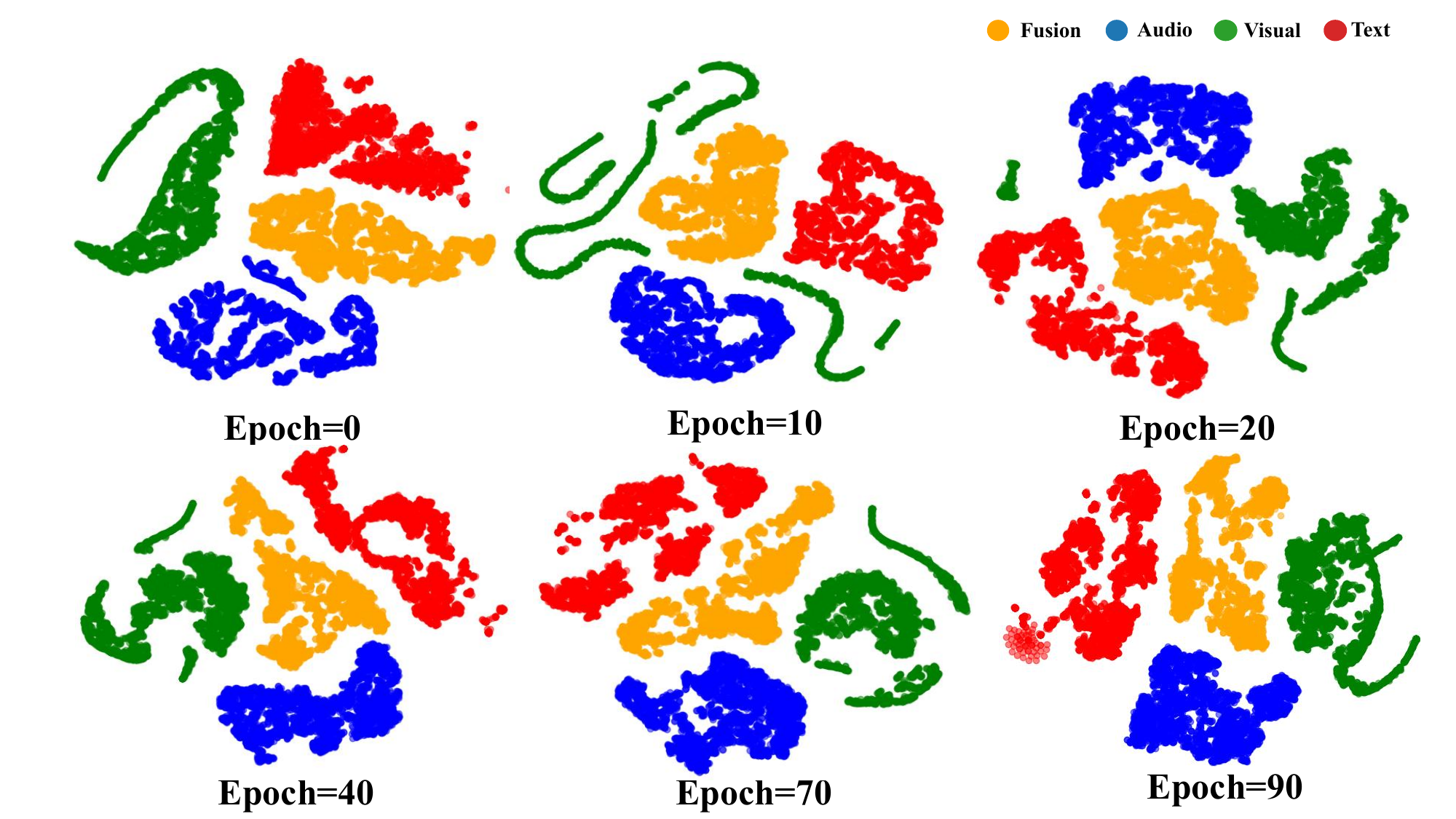}
    \caption{\textbf{Feature distribution analysis.} Evolution of feature distributions over training epochs.}
    \label{fig:feature_distribution}
\end{figure}

\noindent\textbf{Modality Feature Distribution Analysis.}
Figure \ref{fig:feature_distribution} illustrates how feature distributions evolve over training epochs. Initially, before training starts, the visual modality is the weakest, contributing minimally to the fusion features, which are primarily positioned between text and audio distributions. As training progresses, the visual feature distribution gradually shifts, reflecting improved modality balance. By later epochs, the fusion features settle at the center of the three modality distributions, indicating that all modalities contribute effectively. Furthermore, each modality cluster becomes more compact, suggesting that the model learns to generate more cohesive and discriminative representations over time. These results confirm that our adaptive alternating training strategy successfully aligns modality representations and enhances feature fusion quality.

\subsection{Robustness Analysis}
\noindent We conduct robustness analysis on the IEMOCAP dataset, randomly dropping modalities at rates from 0\% to 50\%. Our approach is compared with QMF \cite{zhang2023provable}, MMIN \cite{zhao2021missing}, TATE \cite{zeng2022robust}, and MLA, which are designed for handling missing modalities effectively. As shown in Table \ref{tab:tab4}, our method consistently achieves the best or highly competitive results across all missing modality scenarios. In the complete input setting (0\% missing), it achieves the highest accuracy of 75.07. As the missing rate increases, our model remains robust—reaching 69.56 at 10\% missing and maintaining the best performance even under 50\% missing, where it achieves 53.48. These results demonstrate that our training strategy significantly enhances resilience, allowing the model to preserve high performance despite severe modality degradation.

\begin{table}[t]
\centering
\caption{\textbf{Robustness analysis.} We report fusion accuracy (\%) under different missing modality rates.}
\label{tab:robustness}
\small
\begin{tabular}{l|c|c|c|c|c|c}
\toprule
\textbf{Model} & \textbf{0\%} & \textbf{10\%} & \textbf{20\%} & \textbf{30\%} & \textbf{40\%} & \textbf{50\%} \\
\midrule
QMF & 72.45 & \underline{69.20} & 65.31 & \textbf{63.41} & {57.45} & \underline{53.12} \\
    MMIN & \underline{74.35} & 68.83 & \underline{66.12} & 62.15 & \textbf{58.27} & 51.58  \\
TATE & 72.72 & 68.47 & 65.58 & 61.34 & 55.83 & 51.40 \\
MLA & 72.36 & 52.57 & 52.21 & 47.70& 45.71 & 43.00 \\
Ours & \textbf{75.07} & \textbf{69.56} & \textbf{66.58} & \underline{62.78}& \underline{57.81} &  \textbf{53.48} \\
\bottomrule
\end{tabular}

\label{tab:tab4}
\end{table}

\begin{figure}[!t]
    \centering
    \includegraphics[width=0.9\linewidth]{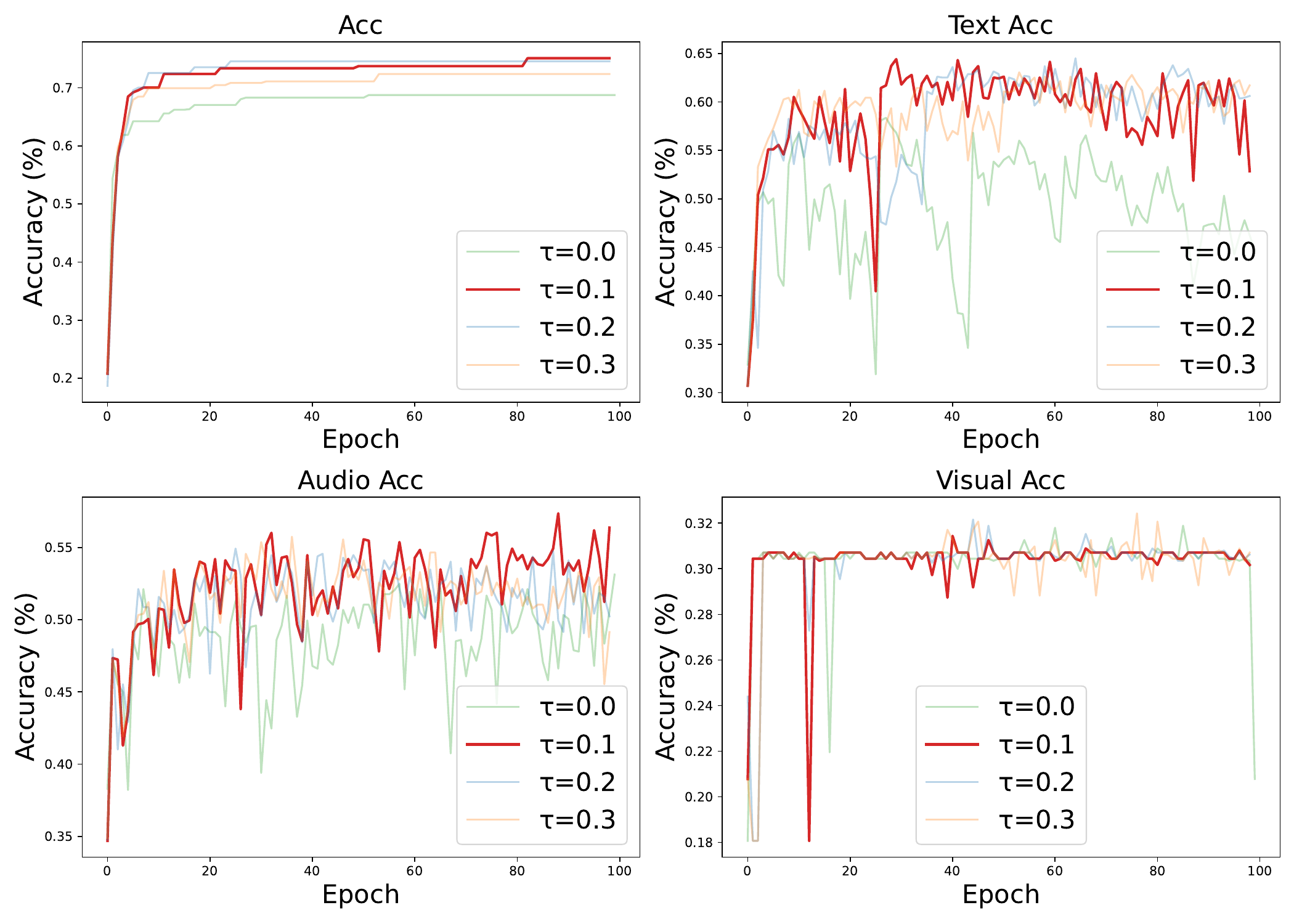}
    \caption{Effect of threshold $\tau$ on modality-specific and fused accuracy on IEMOCAP.}
    \label{fig:threshold}
\end{figure}

\begin{table}[t]
  \centering
  \caption{\textbf{Comparison between fixed threshold $\tau$ and learnable gating for cross-modal interaction regulation on CREMA-D and Kinetics-400.}}
  \small
  \setlength{\tabcolsep}{1mm} 
  \begin{tabular}{lc|cc|cc}
    \toprule
   Data & \diagbox[width=5em]{$Acc$}{$EDM$}& \multicolumn{2}{c|}{$\tau$} & \multicolumn{2}{c}{Gate}\\
    \midrule

   \multirow{2}{*}{CREMA-D}  & Audio & \color{blue}{55.78}&\color{blue}{2.19} & \color{blue}{57.26}& \color{blue}{0.48}\\
                             &Visual & \color{gre}{71.37}&\color{gre}{0.59} & \color{gre}{75.00}& \color{gre}{0.40}\\
                            (ResNet) &\textbf{Multi} & \textbf{84.95}&2.78 & 84.81& \textbf{0.88}\\

    \cmidrule{1-6}
    \multirow{2}{*}{Kinetics-400} &      Audio & \color{blue}{53.21}&\color{blue}{0.20}& \color{blue}{49.96}&\color{blue}{0.66}\\
                            & Visual &\color{gre}{63.67}&\color{gre}{0.11}& \color{gre}{62.90}&\color{gre}{0.40}\\
                           (ResNet) & \textbf{Multi} &79.35&\textbf{0.31}& \textbf{80.21}&1.06\\

    \bottomrule
  \end{tabular}
  \label{tab:gate}
\end{table}

\subsection{Threshold Analysis}

\noindent To regulate cross-modal interactions, we introduce a filtering threshold 
$\tau$ that controls whether cross-modal similarities are retained during alignment. Conceptually, $\tau=0$ represents a minimally restrictive regime where all positively correlated cross-modal interactions are retained, and filtering is applied only to negative or null similarities. As $\tau$ increases, the filtering criterion becomes more selective, progressively removing weak but positive correlations and focusing alignment on higher-confidence cross-modal interactions.

\noindent \textbf{Empirical observation.}  We conduct threshold search by starting from $\tau=0$ and incrementally increasing it with a step size of 0.1. This process reveals a consistent empirical pattern across datasets, as illustrated in Figure \ref{fig:threshold}. When $\tau$ is small, performance remains relatively stable, as weak cross-modal interactions are insufficiently filtered and noise discrimination is limited. As $\tau$ reaches a dataset-dependent critical region ($\tau=0.1$ for IEMOCAP and CREMA-D, and $\tau=0.3$ for Kinetics-400 and MVSA), fusion performance improves noticeably, indicating more effective suppression of spurious alignments. However, further increasing $\tau$ leads to performance degradation, as overly strict filtering removes beneficial cross-modal signals and restricts information exchange.

\noindent \textbf{Learnable gate.} We further introduce a search-free learnable alternative that adaptively regulates cross-modal interactions at the pair level via a lightweight gating mechanism.

For a pair of modality features $\mathbf{M}^{(i)}$ and $\mathbf{M}^{(j)}$, we compute their sample-wise correlation matrix $\mathbf{C}^{(ij)}$ and predict a gating score
\begin{equation}
    g^{(ij)} = \sigma\bigl(f_{\theta}([\mathbf{C}^{(ij)}, \lVert \mathbf{M}^{(i)} \rVert, \lVert \mathbf{M}^{(j)} \rVert])\bigr)
\end{equation}
where $f_{\theta}$ is a small multilayer perceptron. The filtered interaction is then given by $\mathbf{\tilde{C}}^{(ij)} = g^{(ij)} \cdot \mathbf{C}^{(ij)}$, enabling soft and differentiable interaction modulation.

Table~\ref{tab:gate} reports the comparison between the fixed-threshold strategy and the proposed learnable gate on CREMA-D and Kinetics-400. While the learnable gate does not always reach the absolute peak performance achieved by a dataset-specific optimal threshold, it consistently yields competitive multimodal accuracy and exhibits lower sensitivity to dataset characteristics. Notably, on Kinetics-400, the learnable gate even surpasses the best fixed-threshold configuration, indicating its ability to adaptively identify beneficial cross-modal interactions in more complex and large-scale settings. Overall, these results suggest that the gate effectively suppresses noisy interactions while preserving informative cross-modal signals, offering a more robust and scalable alternative to manual threshold selection.

\begin{table*}[t]
\centering
\caption{\textbf{Per-epoch training cost and accuracy comparison on the CREMA-D dataset.}}
\label{tab:efficiency}
\small
\begin{tabular}{l|c|c|c|c|c|c}
\toprule
\textbf{Model} & \textbf{Type} & \textbf{Forward (s)} & \textbf{Backward (s)} & \textbf{Total / Epoch (s)} & \textbf{Relative Cost} & \textbf{Acc}\\
\midrule

Concat & Joint & 12.06 & 23.19 & 35.25& 1.00× & 70.30\\
LFM & Joint & 12.00 & 12.80 & 24.80 & 0.70×  & 81.59\\
MLA & Alternating & 11.80 + 11.82 & 12.19 + 12.21 & 48.02 & 1.36×  & 81.05\\
Ours & Alternating & 10.45 + 10.82 & 12.00 + 25.06 & 58.33 & 1.65×  & \textbf{84.95}\\
\bottomrule
\end{tabular}
\end{table*}

\begin{figure}[!t]
    \centering
    \includegraphics[width=0.8\linewidth]{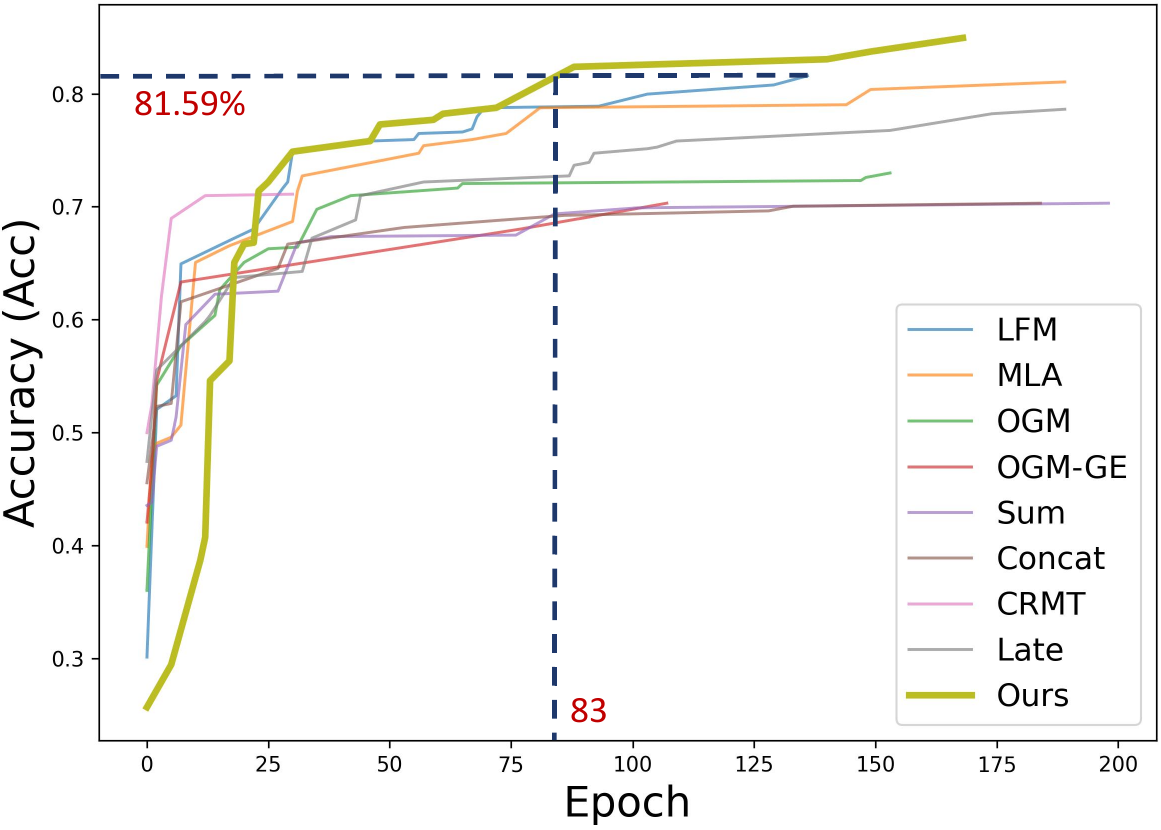}
    \caption{Convergence curves on CREMA-D.}
    \label{fig:convergency}
\end{figure}

\subsection{Training Efficiency Analysis}
\noindent 
To quantitatively assess training efficiency, we conduct a detailed timing analysis on the CREMA-D dataset. We compare four representative models: two joint training baselines (Concat and LFM) and two alternating training approaches (MLA and Ours). All results report averaged wall-clock forward and backward propagation time per epoch.

\noindent 
\textbf{Per-epoch training cost.}
Table~\ref{tab:efficiency} summarizes the per-epoch training time of different methods. The naive joint training baseline Concat requires 12.06s for forward propagation and 23.19s for backward propagation, totaling 35.25s per epoch. The jointly trained LFM model exhibits a lower backward cost, resulting in 24.80s per epoch. For alternating training methods, the actual computational overhead is notably lower than the intuitive expectation of doubling the training cost. The MLA baseline requires 48.02s per epoch, corresponding to only 1.36$\times$ the cost of Concat, as it updates only a subset of parameters at each stage. Our method incurs 58.33s per epoch (1.65$\times$), which remains well below a twofold increase since approximately half of the parameters are updated in the first stage, leading to a substantially reduced backward pass.

\noindent 
\textbf{Computational complexity discussion.}
From a complexity standpoint, alternating training does not alter the asymptotic computational complexity relative to joint training, as all methods share the same backbone architectures. The difference lies in the constant factor introduced by sequential optimization. In practice, this factor is mitigated by partial parameter updates at each stage, significantly reducing backward computation. Compared with MLA, our method introduces additional backward cost only in the second stage by updating an extra subset of parameters, resulting in a moderate increase in per-epoch time while delivering markedly better performance.

\noindent 
\textbf{Efficiency--accuracy trade-off.}
Although our method has a higher per-epoch training cost, it achieves comparable performance in substantially fewer epochs. As shown in Figure~\ref{fig:convergency}, our model reaches the performance level that other methods require more than 125 epochs to attain within approximately 83 epochs. When measured by the wall-clock time required to reach a given performance target, the overall training cost of our method remains within an acceptable range and is significantly lower than the pessimistic estimate suggested by per-epoch cost alone. These results demonstrate a favorable efficiency–accuracy trade-off, indicating that the additional computational overhead introduced by the proposed alternating strategy is both controlled and justified by the resulting performance gains.

\section{Conclusion}
\label{sec:conclusion}

\noindent This work introduces a modality-aware alternating training strategy that dynamically prioritizes weaker modalities to achieve balanced multimodal learning. We further design a simple yet effective memory and mapping mechanism to refine and align representations. Experiments on extensive benchmarks confirm the effectiveness of our method across both standard and LLM-based backbones. In both theory and practice, we demonstrate that the update order plays a pivotal role in alternating training. Specifically, we show that weak-to-strong scheduling enables stronger modalities to absorb and refine earlier errors, thereby driving the model toward a more balanced equilibrium state across modalities. This highlights modality order optimization as a key mechanism for narrowing the gap to balanced multimodal learning.

\bibliographystyle{IEEEtran}
\bibliography{reference.bib}

\clearpage
\appendix
\subsection{Proof of Theorem A.1}

\noindent
To formally prove that weak-to-strong modality training leads to a better convergence bound than strong-to-weak training, we construct a simplified theoretical model and analyze the propagation of memory error and the loss behavior in both sequences.

\subsubsection{Ordered Alternating Optimization.}

Let $M^{(1)}$ and $M^{(2)}$ represent two modalities with distinct representational capacities. Specifically, $M^{(1)}$ is defined as a weaker modality with relatively low expressiveness, while $M^{(2)}$ serves as a stronger modality capable of capturing more informative representations. The fusion loss is defined as:
\begin{equation}
\mathcal{L}_{\text{fusion}}(\theta) = \alpha_1 \mathcal{L}_1(\theta, \mathbf{H}^{(1)}) + \alpha_2 \mathcal{L}_2(\theta, \mathbf{H}^{(2)})
\end{equation}
where $\theta$ denotes the shared parameters involved in the fusion process, and $\mathbf{H}^{(i)}$ represents the memory state maintained during training on modality $M^{(i)}$. The weighting factors $\alpha_1$ and $\alpha_2$ reflect the relative contribution of each modality, with $\alpha_1 \ll \alpha_2$, indicating that the weaker modality $M^{(1)}$ has a lower impact on the final prediction. The memory update follows a weak-to-strong ordering, such that information flows from $\mathbf{H}^{(1)}$ to $\mathbf{H}^{(2)}$. In addition, let $\delta(\mathbf{H})$ denote the error associated with the memory state $\mathbf{H}$, which directly influences the downstream prediction quality.

\subsubsection*{Weak-to-Strong Order: Error Absorption.}

In the weak-to-strong training configuration, modality $M^{(1)}$ is updated first, resulting in a memory state $\mathbf{H}^{(1)}$ that may carry considerable error, denoted as $\delta(\mathbf{H}^{(1)})$, due to its limited representational capacity. Subsequently, the stronger modality $M^{(2)}$ is trained on this intermediate state, allowing it to absorb the noise in $\mathbf{H}^{(1)}$ and refine the shared parameters $\theta$ more effectively. This process exemplifies what we term as \textit{Error Absorption}, where stronger modalities correct earlier imprecisions. The resulting fusion loss under this order can be approximated as:
\begin{equation}
    \mathcal{L}_{\text{fusion}}^{w \rightarrow s} \approx \alpha_1 \delta(\mathbf{H}^{(1)}) + \alpha_2 \cdot \epsilon
\end{equation}

where $\epsilon$ denotes the residual error introduced during the training of $M^{(2)}$, which is significantly smaller than $\delta(\mathbf{H}^{(1)})$, i.e., $\epsilon \ll \delta(\mathbf{H}^{(1)})$.

\subsubsection*{Strong-to-Weak Order: Error Amplification.}

In the strong-to-weak configuration, the training begins with modality $M^{(2)}$, which produces a high-quality memory state $\mathbf{H}^{(2)}$ owing to its stronger representational capacity. This state is then inherited by the weaker modality $M^{(1)}$, which is expected to refine or preserve the fused knowledge. However, due to its limited learning ability, $M^{(1)}$ fails to effectively utilize or maintain the precision of $\mathbf{H}^{(2)}$, resulting in degraded memory quality. This phenomenon exemplifies what we term as \textit{Error Amplification}, where early accurate representations are corrupted by subsequent weaker updates. The resulting fusion loss in this setting is approximated by:
\begin{equation}
\mathcal{L}_{\text{fusion}}^{s \rightarrow w} \approx \alpha_2 \cdot \epsilon + \alpha_1 \cdot \delta(\mathbf{H}^{(2)})
\end{equation}
where $\epsilon$ represents the residual loss during the initial strong modality training. Since the weak modality cannot absorb or preserve the previously learned information, the propagated error $\delta(\mathbf{H}^{(2)})$ remains large—often approximating or exceeding $\delta(\mathbf{H}^{(1)})$.

\subsubsection*{Comparative Gap Analysis.}

To quantify the effectiveness of different training orders, we define the fusion loss gap between strong-to-weak and weak-to-strong configurations as:
\begin{equation}
\Delta \mathcal{L} = \mathcal{L}_{\text{fusion}}^{s \rightarrow w} - \mathcal{L}_{\text{fusion}}^{w \rightarrow s} = \alpha_1 \left[ \delta(\mathbf{H}^{(2)}) - \delta(\mathbf{H}^{(1)}) \right]
\end{equation}
where $\delta(\mathbf{H}^{(i)})$ denotes the error in the memory state of modality $M^{(i)}$. Since the weak modality cannot effectively absorb or preserve the representation from the strong one, we have $\delta(\mathbf{H}^{(2)}) \ge \delta(\mathbf{H}^{(1)})$, implying
\begin{equation}
\Delta \mathcal{L} > 0 \quad \Rightarrow \quad \mathcal{L}_{\text{fusion}}^{w \rightarrow s} < \mathcal{L}_{\text{fusion}}^{s \rightarrow w}.
\end{equation}
This inequality holds under the following assumptions: (1) the weak modality contributes less to the overall loss, i.e., $\alpha_1 \ll \alpha_2$; (2) errors from weak modalities can be effectively absorbed by stronger ones; and (3) a weak-to-strong update order facilitates such error absorption. These conditions collectively lead to the conclusion that
\textbf{weak-to-strong training provides a provably tighter upper bound on fusion loss},
establishing it as a theoretically preferred strategy over strong-to-weak or unordered updates.

\begin{figure}[!t]
    \centering
    \includegraphics[width=\linewidth]{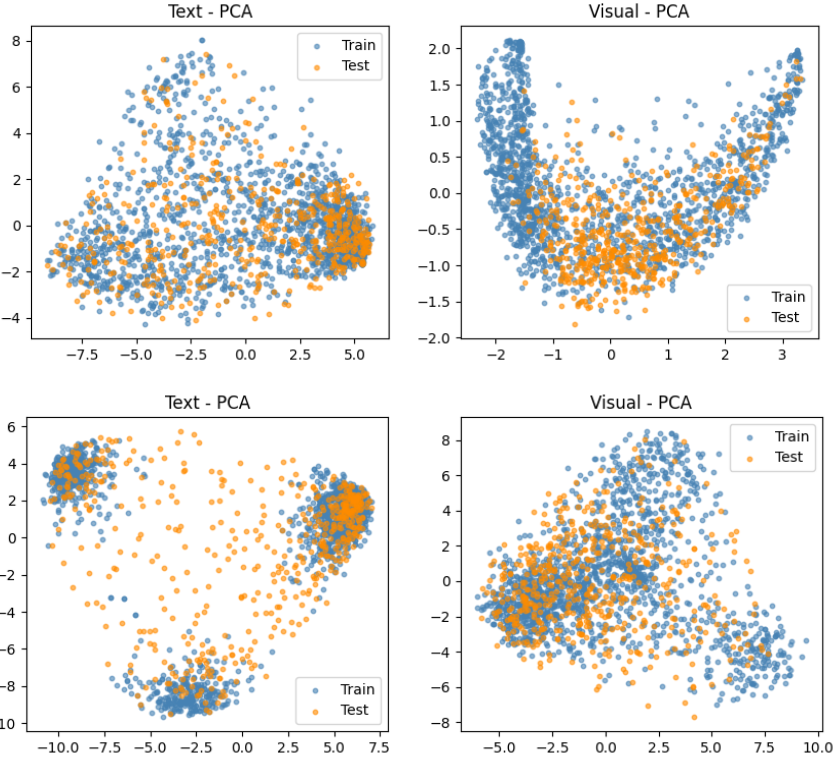}
    \caption{PCA distributions of different modality features before training (top) and after training (bottom).}
    \label{fig:pca}
\end{figure}

\subsection{Data Distribution Analysis}
To evaluate the distributional consistency assumption that underpins our memory-guided inference strategy, we conduct a principal component analysis (PCA) on the MVSA dataset to visualize the embedding distributions of training and test samples under both textual and visual encoders.

The results, illustrated in Figure~\ref{fig:pca}, reveal notable shifts in representation distributions before and after training, with particularly pronounced discrepancies across modalities. These discrepancies reveal that the fixed weak-to-strong scheduling, while theoretically grounded, may not fully accommodate the distributional shifts observed at test time. This gap indicates that current performance still falls short of the theoretical upper limit, implying opportunities for improvement via more adaptive inference strategies.

\begin{figure}[!t]
    \centering
    \includegraphics[width=\linewidth]{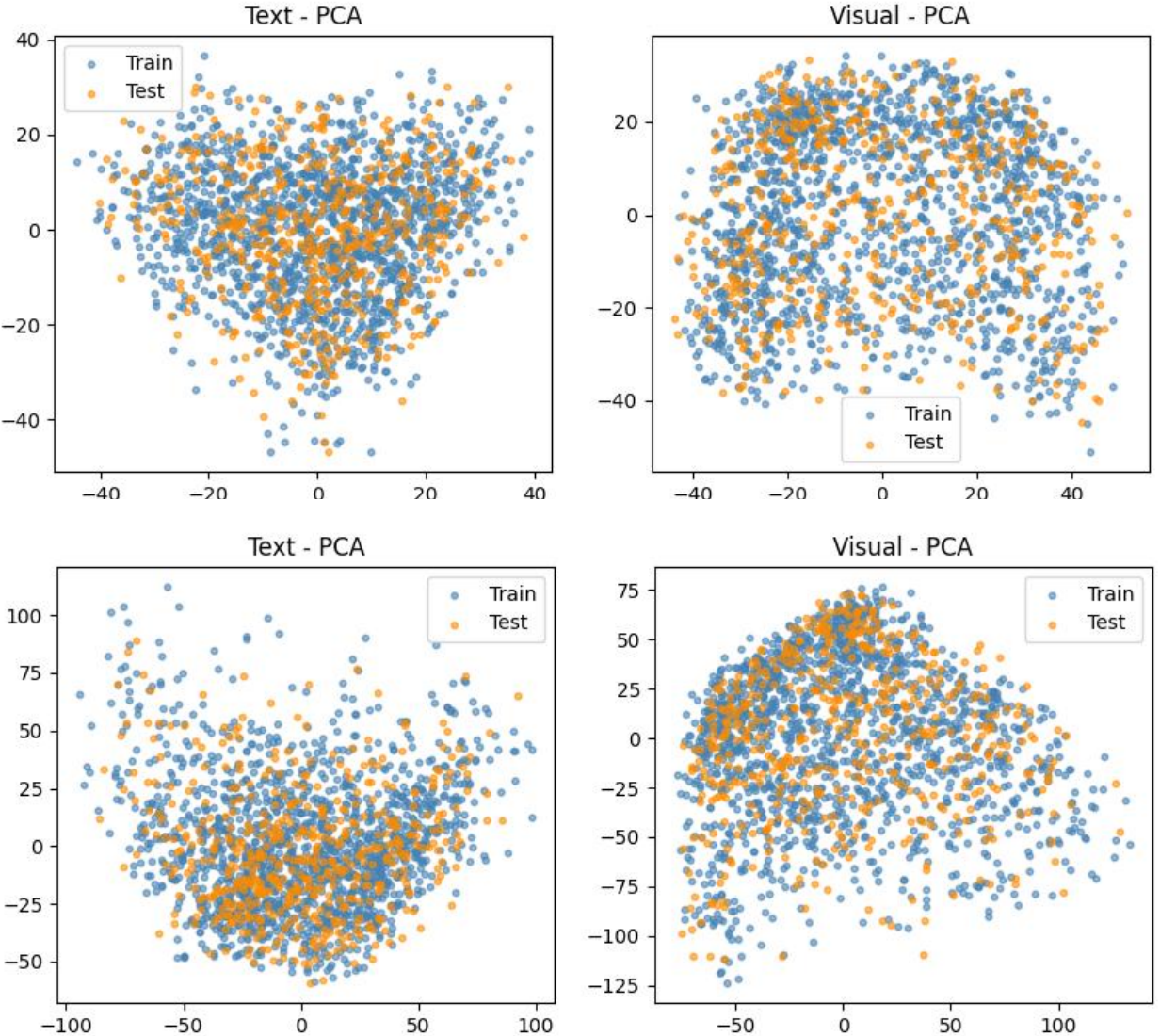}
    \caption{PCA visualizations of modality features extracted by MLLM before (top) and after (bottom) training.}
    \label{fig:pca_mllm}
\end{figure}

\subsection{Encoder Capacity Analysis}
In addition to data distribution, the representational capacity of modality-specific encoders fundamentally influences the model's performance ceiling. As shown in Figure~\ref{fig:pca_mllm}, adopting MLLM-based encoders (e.g., Ola-7B) leads to more consistent cross-modal feature distributions and balanced contributions between textual and visual modalities. This alignment yields notable performance improvements over traditional unimodal encoders, as evidenced in Table~\ref{tab:tab1}, even under distributional shifts.

These results indicate that stronger encoders can inherently reduce the impact of distribution mismatches by enhancing cross-modal consistency and bringing estimated modality importance closer to its true value. As a result, overall performance approaches the theoretical upper bound more closely, underscoring the joint influence of data distribution and encoder capacity in multimodal learning.

\begin{figure}[!t]
    \centering
    \subfloat[\footnotesize MVSA]{
        \centering
        \includegraphics[width=0.48\linewidth]{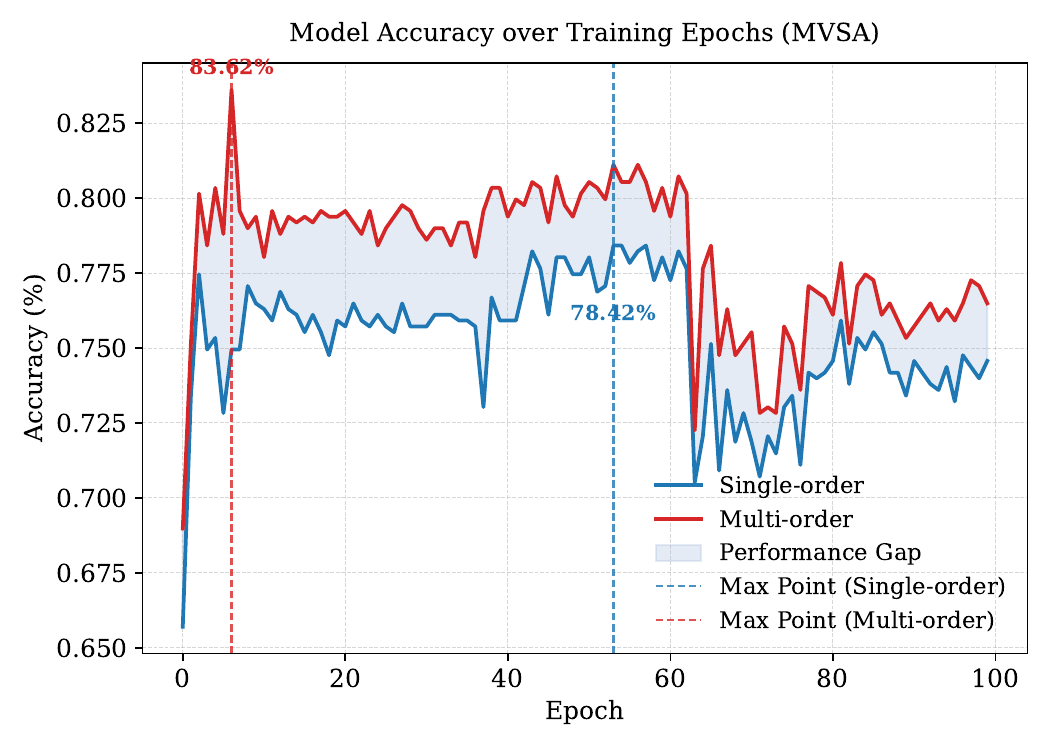}
        }
    \subfloat[\footnotesize CREMA-D]{
        \centering
        \includegraphics[width=0.48\linewidth]{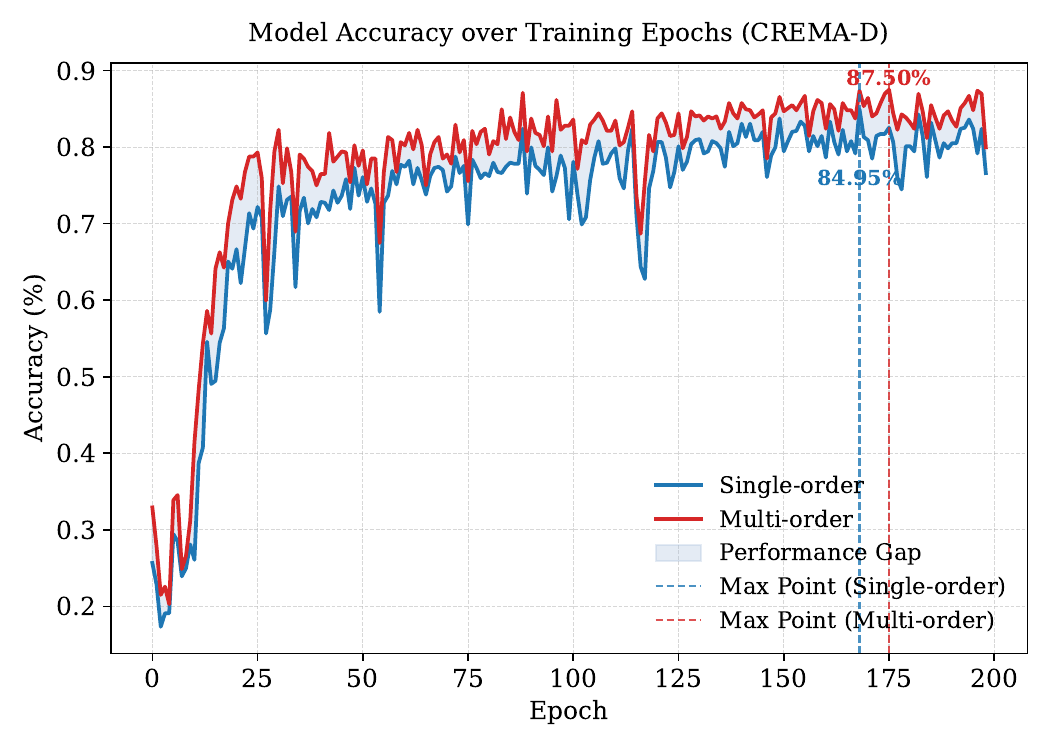}
        }
        
    \subfloat[\footnotesize Kinetics-400]{
        \centering
        \includegraphics[width=0.48\linewidth]{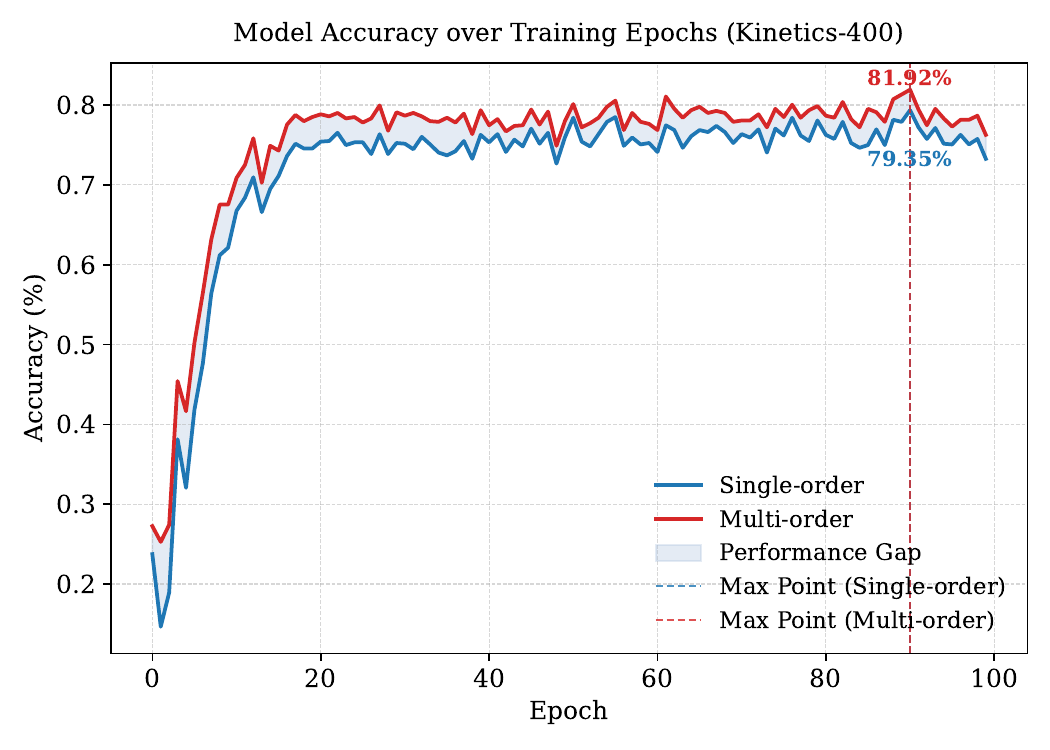}
        }
    \subfloat[\footnotesize IEMOCAP]{
        \centering
        \includegraphics[width=0.48\linewidth]{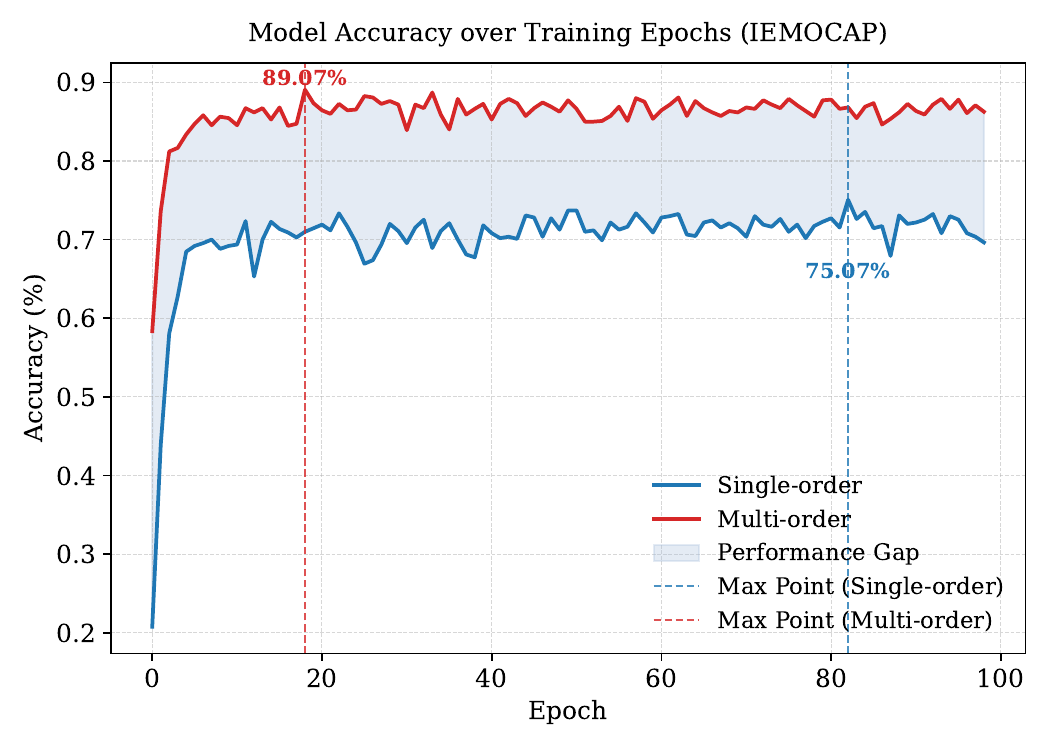}
        }
    \caption{Modality Capability Upper Bound vs. Actual Performance on Four Datasets.}
    \label{fig:theoretical_performance}
\end{figure}

\subsection{Performance Upper Bound}
To empirically evaluate the theoretical performance ceiling of our method, we compare the actual performance trajectory with an idealized upper bound across four benchmark datasets. The upper bound is estimated under oracle conditions, where modality scheduling is perfectly aligned with the true relevance of each modality for every test instance.

As illustrated in Figure~\ref{fig:theoretical_performance}, the performance gap is most prominent in MVSA and IEMOCAP, with up to 5\%–\% difference between observed and ideal performance due to distributional shifts and modality imbalance. In contrast, datasets like CREMA-D exhibit a narrower gap (around 3\%), reflecting more consistent modality contributions and balanced encoder strength. These findings demonstrate that while our memory-guided fusion strategy is effective, its performance is still bounded by dataset properties and encoder capacity. This highlights a promising direction for future work—adapting inference-time modality integration order to better match test-time modality relevance and approach the theoretical limit.

\begin{table}[!t]
\small
\centering
\caption{Analysis of fusion strategies on MLLMs.}
\begin{tabular}{@{}l|lll|lll@{}}
\toprule
& \multicolumn{3}{c|}{MVSA-Single}  &  \multicolumn{3}{c}{CREMA-D$^\dag$}    \\ \midrule
   & Text & Visual & Multi & Audio & Visual& Multi  \\ \midrule
Joint & 64.74&  50.48&  73.60& 23.39 & 16.94 &   51.61 \\
Sum & 70.33&  59.15&  78.81& 34.68 & 16.13 & {54.84}\\
Concat & 73.41 & 62.24 & {79.96} & 42.74 &{23.39} & {54.84}  \\ \midrule
\end{tabular}

\label{tab:tab4}
\end{table}

\subsection{Fusion Strategies on MLLMs}

\noindent We conduct experiments on \textbf{MVSA-Single} and \textbf{CREMA-D}, using Ola-7B as the backbone. To evaluate the generalization ability under data-limited scenarios, only 15\% of the CREMA-D training set is used. As a baseline specific to MLLM architectures, we compare our alternating training framework with \textbf{Joint}, a commonly used fine-tuning paradigm where modality-specific features are concatenated and fed into the MLLM simultaneously for classification. In this setting, the MLLM serves as the fusion backbone. While this joint approach benefits from the MLLM's intrinsic cross-modal alignment, our results (see Table \ref{tab:tab4}) show that it consistently underperforms compared to simple fusion strategies such as Sum and Concat. This indicates that the inherent alignment capability of MLLMs is still insufficient for fully capturing modality interactions, and that explicitly modeling modality-wise optimization through alternating training provides a more effective learning paradigm.

\subsection{Computing Infrastructure}
Table~\ref{tab:infrastructure} summarizes the primary computing infrastructure employed in our experiments. Additional software dependencies are provided in the source code (requirements.txt).

\begin{table}[t]
\centering
\caption{Computing infrastructure used in our experiments.}
\label{tab:infrastructure}
\begin{tabular}{p{0.35\linewidth} p{0.60\linewidth}}
\toprule
\textbf{Component} & \textbf{Specification} \\
\midrule
GPU & NVIDIA H100 80GB $\times$ 1 \\
CPU & Intel(R) Xeon(R) Silver 4114 @ 2.20GHz \\
System Memory & 125 GB \\
Operating System & Ubuntu 22.04.5 LTS \\
Python & Python 3.10.16 \\
PyTorch & Torch 2.5.1 \\
Transformers & Transformers 4.48.0 \\
\bottomrule
\end{tabular}
\end{table}
\end{document}